# Data-driven deep density estimation

Patrik Puchert[1] · Pedro Hermosilla[1] · Tobias Ritschel[2] · Timo Ropinski[1,3]

**Abstract**
Density estimation plays a crucial role in many data analysis tasks, as it infers a continuous probability density function (PDF) from discrete samples. Thus, it is used in tasks as diverse as analyzing population data, spatial locations in 2D sensor readings, or reconstructing scenes from 3D scans. In this paper, we introduce a learned, data-driven deep density estimation (DDE) to infer PDFs in an accurate and efficient manner, while being independent of domain dimensionality or sample size. Furthermore, we do not require access to the original PDF during estimation, neither in parametric form, nor as priors, or in the form of many samples. This is enabled by training an unstructured convolutional neural network on an infinite stream of synthetic PDFs, as unbound amounts of synthetic training data generalize better across a deck of natural PDFs than any natural finite training data will do. Thus, we hope that our publicly available DDE method will be beneficial in many areas of data analysis, where continuous models are to be estimated from discrete observations.

**Keywords** Density estimation · Deep learning · Data-driven · Kernel density estimation · Probability density function

## 1 Introduction

Many data analysis problems, reaching from population analysis to computer vision [6, 28], require estimating continuous models from discrete samples. Formally, this is the density estimation problem, where, given a sample $\{x_i\} \sim p(x)$, we would like to estimate the probability density function (PDF) $p(x)$. Our aim is to enable this at high speed and quality with very little assumptions about the data $\{x_i\}$ or the PDF $p(x)$. While this is a well-solved problem for PDFs on a 1-dimensional (1D) domain, it becomes increasingly difficult in higher-dimensional domains and additionally implies a strong bias on the estimate. Practically limiting is also the fact that sophisticated estimators require long computing times. To achieve the task of finding a good and fast estimator for arbitrary domain dimensions and to make as little assumptions on the PDF as possible (low bias), we employ deep learning, which has recently been successfully applied to many real-life applications in different fields [3, 12, 33].

Although it has been proposed more than 60 years ago, kernel density estimation (KDE) [27, 29] is today still the method of choice when performing density estimation. Unfortunately, a key problem for KDE is the choice of the required *bandwidth* parameter. While automatic approaches exist (such as Silverman [34]), they only work under given assumptions, struggle with computational efficiency and fail for cases where there is no constant bandwidth to explain the entire sample. As a consequence, various neural density estimation approaches have been proposed. Unfortunately, these either require to be trained on the distribution to be analyzed [2, 24], which is not known in most cases, or they have problems generalizing, as they are trained on a single sample only [14, 25, 26].

In contrast to previous work, our method allows for accurate, instantaneous estimation from only a single sample of an arbitrary number of data points or dimensions, without access to the original PDF during estimation, neither in parametric form nor in the form of many samples. To do so, we do not make any assumptions through priors and do not require re-training of the model on the sample at hand during estimation.

✉ Patrik Puchert
  patrik.puchert@uni-ulm.de

1  Institute of Media Informatics, Ulm University, James-Franck-Ring, Ulm 89081, Germany

2  Department of Computer Science, University College London, 66-72 Gower Street, London WC1E 6EA, UK

3  Department of Science and Technology, Linköping University, Linköping 58183, Sweden

To propose an automated and generally applicable neural network approach for direct density estimation, two contributions need to be made. First, a representative training dataset needs to be generated. And second, an appropriate neural network architecture needs to be designed. When reflecting on these two contributions, it becomes clear that they are tightly intervened. The question here is, how to generate a representative dataset, which results in a general applicability of the trained model. The key enabling idea here is to use convolutions on the unstructured distribution to allow for a generalization on arbitrarily sized samples. This restricts the input of the network for each pointwise density estimation to only a finite receptive field. Thus, the training data do not need to represent all possible distributions, which would be infeasible, but rather all possible local characteristics. This is not only a much easier task, but its validity for accurate density estimation has also been shown by the analysis on nearest neighbor estimation [8].

To incorporate these local features, we handle our samples as follows. For any sample $S \subset \mathbb{R}^d$ the receptive field is given by the distribution of the $k$ closest points in the proximity of arbitrary query points $x \in \mathbb{R}^d$. Since there is a limited variety of these structures, a representative training dataset can be generated, which eventually enables generalization to any given distribution. Together with a specifically designed network architecture, we will show that such a data-driven approach can be used for general deep density estimation. To this end, we make the following key contributions:

- Generation of general probability distributions with ground truth PDFs for arbitrary domain dimensions, reflecting the widest possible range of local characteristics.
- Accurate density estimation with a deep neural network in inference, by utilizing a novel convolutional architecture.

In the remainder of this paper, we will first discuss prior work, before detailing our neural network architecture design and the training data generation process. Finally, we will evaluate the proposed method with respect to result accuracy and its generalization capabilities and conclude by summarizing the obtained results.

## 2 Related work

As the density estimation problem is crucial for many data analysis tasks, a substantial amount of prior work has been dedicated toward its solution.

### 2.1 Conventional density estimation

Basic models fit parametric functions such as lines or Gaussians to the sample, but real data are typically of a more complex shape and mixtures of parametric curves are required [9]. Another straightforward solution is to construct discrete histograms with finite-sized bins around centers $c_i$ and counting how many sample points fall into each bin. A high count corresponds to a high value $p(c_i)$. This does not scale well with the number of dimensions, as it requires exponential memory, does not produce a continuous result as only discrete centers are domained, and choosing the bin size can be difficult [32].

Histogram binning is a special case of KDE, where the hard counting is replaced by a compact and soft (e.g., Gaussian or Epanechnikov) kernel $K(||x_i - x||/h)$ that weights the contribution of sample point $x_i$ to a continuous coordinate $x$. KDE has one control parameter $h$, called the *bandwidth* of the kernel. Choosing $h$ can be difficult: if it is too small the result is noisy; if it is too large the result may turn out overly smooth. Picking the optimal $h$ is easy if $p$ is known, but typically we do not have access to $p$. In practice, the easiest way to choose $h$ is based on heuristics due to Silverman [34] that account for the number of points and the domain dimension. More sophisticated methods were suggested to pick the right bandwidth when making prior assumptions on $p$ such as smoothness or sparsity [17]. These are however only applicable if $p$ fits the assumptions, they struggle with computational efficiency and fail for cases where there is no constant bandwidth $h$ to explain the entire sample. While the time complexity of vanilla KDE is only $\mathcal{O}(n)$, it can be as large as $\mathcal{O}(n^2)$ for a KDE with sophisticated bandwidth selection algorithms [38]. A more sophisticated method for density estimation is the nonparametric PDF estimator by Farmer and Jacobs [13]. While it can be applied without the need of fine-tuning parameters such as the bandwidth in KDE and produces good results, it is only applicable for 1D distributions.

### 2.2 Neural density estimation

The most common approach for neural density estimation is the utilization of Gaussian mixture models [10] or their addition the variational Bayesian Gaussian mixture models [4]. However, they require large distributions in order to estimate complex PDFs well enough. Also neural networks have been used to model density estimation for cases where the PDFs $p$ is known and can be used during supervision [2, 24]. Regrettably, we rarely know the PDF of most natural signals: no supervision is available when for instance discovering novel dynamics, group formations, or particle events. Consequently, these methods are limited to estimate density if the PDF is known. While such models are also applied without access to the true PDF [23], the training target for these cases is just generated using KDE or other density estimators, which essentially just shifts the problem from estimating densities to learning KDE, which will hardly ever produce better results than KDE can.

Unsupervised methods that only require access to the sample but not to the PDF, have also been proposed [14, 25, 26]. Here, the network itself represents the PDF, and—while more complex than a linear, Gaussian or parametric mixture model—this essentially is a fit of a network to the data in one sample. This means a new network is trained for every sample. Fortunately, this makes little assumptions, but regrettably, at the expense of quality as it fails to capture the essence of learning: representing previous knowledge relevant for a task; nothing discovered in one sample will ever contribute to density estimation of another sample.

## 3 The DDE method

DDE solely uses local information in the distribution around a query point, to predict an accurate density estimate, whereby a network is used that is trained independently of the query distribution. The core of the DDE approach is a neural network in form of a multi-layer perceptron (MLP), convolved over the complete input. The most important difference to all prior approaches is that we designed the method such that only a single training on a large and representative set of samples with known PDF $p$ is required. The thus-trained model can then be used without the necessity of further training, to estimate the probabilities of arbitrary unknown distributions residing in a domain with same dimensionality.

### 3.1 Network architecture

The architecture of our DDE approach is illustrated in Fig. 1. The network takes as input the sample distribution $S \subset \mathcal{S}$ for $\mathcal{S} = \prod_{i=1}^{d}[a_i, b_i]$ with $a_i, b_i \in \mathbb{R}$ and $a_i \leq b_i$ and a set of query points $Q \subset \mathcal{S}$. For the direct density estimation of each point in the distribution, both sets are the same. The first step is then the realization of a finite receptive field by calculating the distances to the $k$ closest non-identical points $x_i \in S$ to any query point $x \in Q$ by either using a kd-tree, ball tree, or brute force. The latter is only beneficial because of the possibility of full parallelization on a GPU (graphics processing unit). This first step is what makes the network convolutional, as it convolves over the input with the size of the convolution determined by $k$. To achieve a model which can be applied to arbitrary datasets, the convolution is of uttermost importance. While a convolution could be realized by feeding the unstructured points by position, without a direct distance encoding, this would require at least a fine-tuning of the model architecture for any given dimensionality, as the complexity of the input will increase with dimensions.

The obtained distances are then fed into an MLP which predicts $p(x)$. As the MLP is shared over all points, this becomes effectively a convolution with the MLP as convolution kernel and the kernel size defined by $k$. The specific layout of the MLP, i.e., its number and size of hidden layers, is not necessarily fixed but can be determined during each training session, where the resulting estimator is the best model out of an ensemble of trained models. While the value of $k$ could give a bias on the estimate, with highly fluctuating values for small $k$ and overly smoothed estimates for large $k$, we have found empirically that a value of $k = 128$ gives highly accurate results on a wide range of dimensionalities and sample sizes. The evaluation on the choice of $k$ is discussed in Sect. 3.3. Thus, the most important difference to all prior approaches is that we designed the method such that it does not fit the model to $S$ at hand, but uses only the learned information of the training set to predict $p$ for $S$. While it would be possible to assign further one-dimensional convolutions on the distance input, we have found empirically that the MLP achieves the better results. For the exact architecture of the MLP we investigated three types of structures. The first started with an input layer with 2048, 1024, 512 or 256 nodes with every consecutive layer being

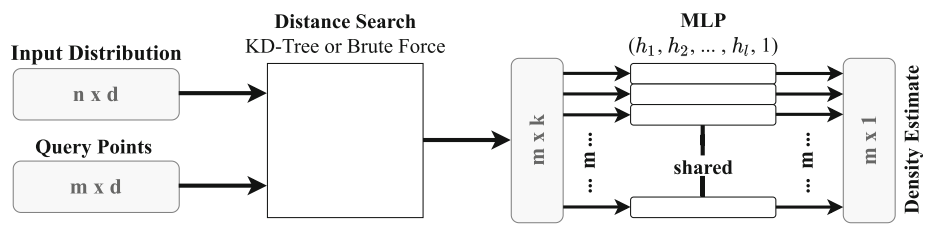

**Fig. 1** Our DDE approach: input is the distribution of $n$ data points with a domain dimensionality $d$, as well as the set of $m$ points to query for density estimation with same dimensionality. For the case of direct density estimation of the distribution, both sets are the same.

Using a kd-tree or a brute force algorithm the distances to the $k$ nearest neighbors are computed. These are fed per query point into an MLP which returns the scalar density estimate for that point equally large or smaller. The second started with an input layer of size 32, 64, 128 or 256, which extends to a hidden layer with a maximal size of 512, 1024 or 2048 nodes, before reducing to the output size of 1. The last type is similar to the first one, but uses a skip connection between every two consecutive layers, inspired by ResNet [16]. All of these models were tested with different numbers of hidden layers. Testing these along different sizes for $k$ we have achieved the best results using the second type of models with $k = 128$, 128 input neurons, 9 hidden layers with a maximum size of 512 and a doubling/halving of layer sizes, respectively, between each two consecutive layers.

This architecture results in a time complexity for DDE of $\mathcal{O}(n \log n)$, which is governed by the distance search. While the calculations from DDE to get the density estimate from distances are implemented in TensorFlow, and thus highly parallelized on the GPU, the time-wise most complex part, i.e., the distance search, is still implemented on the CPU (central processing unit).

## 3.2 Training process

For the training objective we used the mean squared error as loss function. Apart from that, we used conventional building blocks, whereby our model was trained with rectified linear units (RELU) [19] as activation function in every layer, with batch normalization [18] after every but the last layer, and to calculate the parameter update we used the Adam optimizer [21] with default parameters and exponential learning rate decay.

The output of the model in 1D is further smoothed by a univariate spline interpolation with a fixed order of the polynomial and an automated adaptive smoothing factor, making this amendment also fully automated. Such post-processing is not conducted in higher dimensions, as we know of no existing method which can be automated, allowing for an unbiased way to return smoother but still accurate estimates over all applications.

## 3.3 On the choice of $k$

The parameter for the number of neighbors fed into the network $k$ was empirically found and set. Here, we briefly discuss how we determined $k = 128$. As the choice of the parameter $k$ is intertwined with the network architecture, especially the size of the first hidden layers, we present here only an example of the empirical analysis we conducted. This can be seen in Fig. 2, where we plot the Kullback–Leibler (KL) divergence against the parameter $k$ for 1D and 3D. Note that these models show slightly poorer estimates in 1D and even poorer ones in 3D than the ones discussed in the rest of this paper, as they were trained much shorter, which also leads to the rather large noise regarding different $k$. The model trained here differed only in the number of $k$ for the convolutional window, but had the same architecture otherwise, with the first hidden layer of size 128. The significant performance loss for $k \gtrsim 135$ 1D is attributed to this fact, as the lower size of the first hidden layer with respect to the input needs a fast information encoding already in the first layer, while for hidden layers larger than the input, the information can be "passed through" the first layers and be iteratively encoded by the entire network. While the former should be in general possible, in practice the model will be caught too quickly in a local optimum for the first layer. For our investigations we tested a large variety of different architectures with different layer sizes and numbers, which description would be beyond the scope of this paper. For the same reason our tests did not include densely sampled $k$, but such from a power 2 distribution (i.e., 16, 32, 64, 128, 256). Having that said, the final selection for $k$ involved mainly the error metrics over the validation dataset, but as well the visual quality of the resulting estimates and the fact that $k$ should not be too large, as it sets an ultimate bound on the lowest sample size possible to estimate. Regarding the error metrics we show the KL divergence in Fig. 2 on the top, where we can see first of a strong performance gain for larger $k$ which asymptotically reaches a minimum for $k \gtrsim 60$, and as discussed above becomes worse again for $k \gtrsim 135$. This indicates that $k$ could be selected as low as 64 according to the larger test series. In the next step, we also assessed the visual quality of the estimates (Fig. 2, bottom). We have analyzed unsmoothed estimates for $k = 32$ (*blue*), for $k = 64$ (*orange*) and for $k = 128$ (*green*) in comparison with the true PDF (*black*), for 2 sample distributions of size 1000. The reason for choosing $k = 128$ over $k = 64$ can be seen in the lower sensitivity to the noise in the data; thus, the estimate for $k = 128$ is smoother. An explanation to the empirically found constant optimal value for different dimensions is the fact that the model is trained for every dimensionality and takes only the scalar distance as input. Thus any dimension-specific dependency of the distances is inherently encoded by the network.

An additional problem of our estimates becomes visible in Fig. 2 (*bottom*), where we can see a false zero estimate on the tail of the distributions. Solving this problem requires the longer training times as well as the selection of the final model from a larger set of identical models, where the last step ensures that we rule out models which converged to bad local optima. While this seems like a rather time-intensive training, it should be noted that it is quickly surpassed by the generation of the training data, and that the model has to be trained only once for all future applications on the same domain dimensionality.

**Fig. 2** Comparison of different values for *k* of the DDE model. The KL divergence is shown as mean over the validation dataset with 250 functions, plotted against the number *k* in 1D (*top*) and 3D (*center*). On the bottom two examples of the validation set in 1D are shown, each with a size of the distribution of 1000 samples, with the model estimates for $k = 32$, $k = 64$ and $k = 128$, respectively

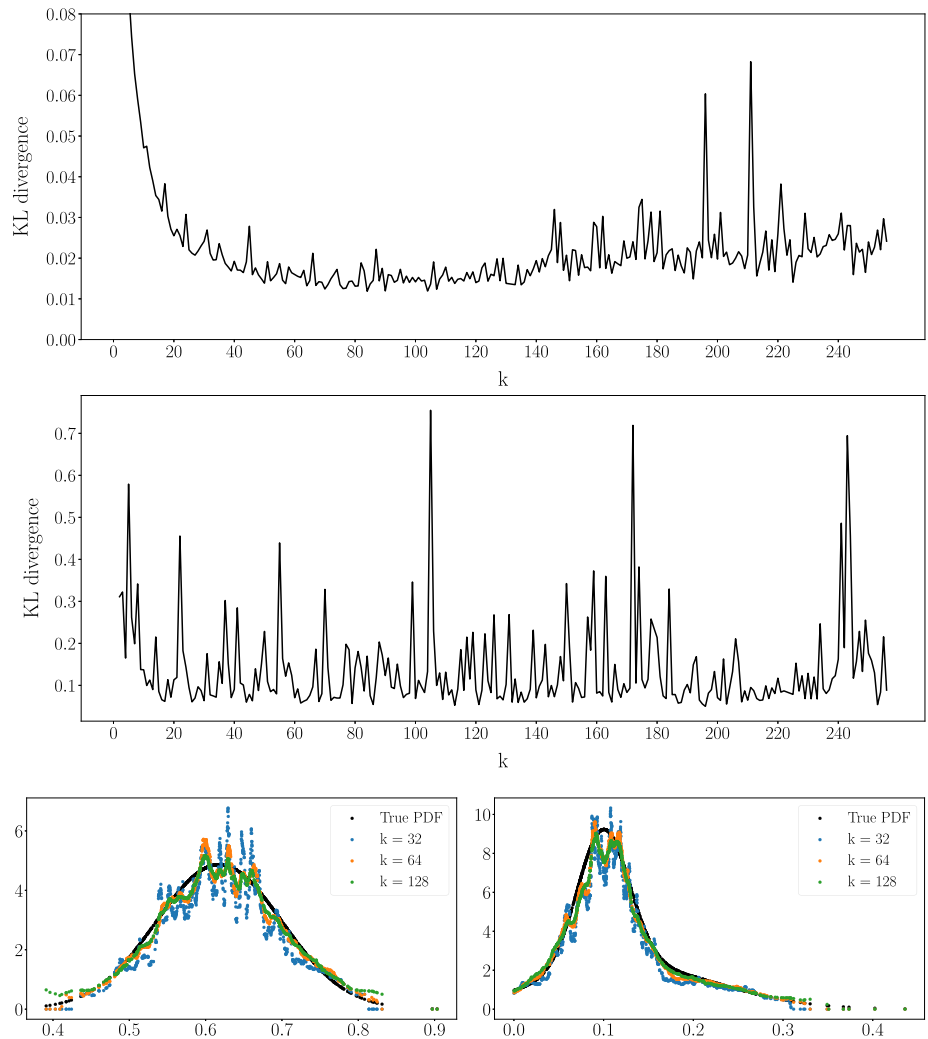

# 4 Data generation

Besides the local learning concepts described above, an appropriate dataset is crucial in order to realize DDE. The samples used to train our network are purely synthetic. This is necessary as a great amount of data along with its ground truth, i.e., the actual PDF values at every sample point, is required during training, but also when evaluating the trained models and comparing the results to state-of-the-art approaches. Additionally, the training data must supply a wide as possible range in the feature space, regarding the structures of PDFs.

To obtain such a representative training dataset, we propose a simple algorithm which generates the desired number of probability distributions by selecting functions from a set of 1D base functions and connecting them with a random operator from a set of defined operators. To achieve greater randomness in the resulting functions we equipped the base functions with random factors in various places, combined this with a varying extent of the domain, which is scaled to unit range [0, 1] afterward. Examples of such synthetic 1D functions are presented in Fig. 3 We note that applying the network on these scaled data does in general not prohibit estimations outside of this range, as it is fed with distances, which can still be obtained for such data points. Of course this would change for distances outside of the trained range, i.e., larger than 1, but such data points would either be in the original scaled distribution if they have a notable probability, or have vanishing probability $p(x) \rightarrow 0$, and are thus not respected by our algorithm. While the different kinds of base functions themselves, applied in this manner, define the local structure of the obtained PDFs, randomization of the base function with respect to its relative position on the *x*-axis and the randomization of the domain extent of the base function are important to cover a greater portion of the feature space with respect to the global PDF structure. To frame this in an example, we for instance use a Gaussian base function and apply DDE. As DDE predicts the PDF estimate only from distances, the position of the base function on the $x$-axis makes no difference. However, combining this first base function with the positive part of a sine by addition, shows the importance of both randomization aspects. The randomization of the position of the Gaussian in a $\pi$-range does not drastically change the global structure of the resulting PDF, while the randomization of domain extent does not lead to the appearance or disappearance of periodic features in the PDF, caused by the sine.

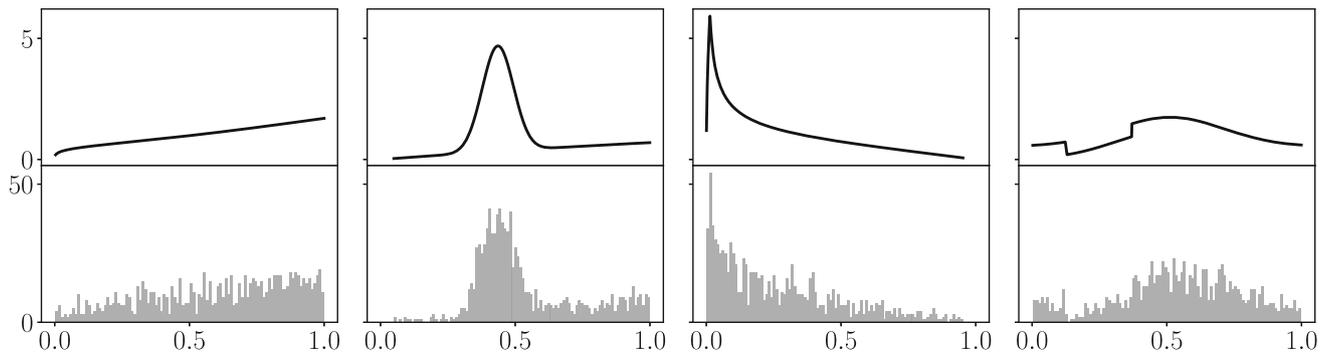

**Fig. 3** Example PDFs (*top*) with associated histograms (*bottom*) from the synthetic training set used for 1D PDF prediction

For an even greater randomization, also the number of functions connected $n_c$ is varied during function generation. The set of functions was selected, such that they contain a wide range of different structures. These are periodic or aperiodic features, (non-)monotonicity, different degrees of slopes, signal peaks of varying degree, discontinuities, valleys, (non-)heavy edges, and semi-diverging features. The latter needs to be semi-diverging, as actually diverging functions lead to problems in the PDF generation because of exploding values. For further details on the set of base functions we refer to Appendix 2.

For the set of operators in this work we used only the addition and multiplication to guarantee the positive definiteness of the synthesized functions, since a cutoff used on non-positive definite functions as well as a power operator, albeit positive definite, would result in too many redundant features in the resulting functions.

While it would be optimal to construct PDFs for larger domain dimensions $d > 1$, it would become an increasingly tedious task to find a proper set of base functions, which would have to be done for every given domain dimensionality. Instead, we adapt our method to always select functions from the same set of 1D base functions as before and combine them to higher functions with a higher-dimensional domain. This is realized through two different approaches. In one approach first $n_c$ base functions are coupled to obtain $d$ 1D functions which are then coupled to $d$-dimensional functions. The other approach instead first constructs $n_c$ $d$-dimensional functions from the 1D functions and couples them afterward with a randomly selected operator, which is in practice again either addition or multiplication. The functions built in this way are normalized by numerical integration to obtain PDFs, and probability-distributed samples are drawn from these via rejection sampling. The benefit of the former approach is that it is significantly faster to construct as small as possible upper bound for the importance sampling and to normalize the function, as the dimensions are decoupled (linear growth with the number of dimensions opposed to an exponential growth), but it bears the drawback that the functional space is more structured and thus the feature space is less covered. Both generation schemes still pose the problem of exploding or vanishing numerics for large dimensions. In detail, this arises for base functions which have a very small maximum. To counter this it is necessary to apply additional constraints on the high dimensional function generation. Therefore, starting at $d = 50$, first the set of operators for combining the base functions is reduced to only the addition. Secondly, we neglect such base functions which have a maximum lower than 0.01. Otherwise the small values of the functions, especially paired with a multiplication operator, will lead to exploding values during normalization of the function, to obtain a PDF. To ensure our experiments are feasible, we limited the number of dimensions and employed $2 \leq n_c \leq 7$ for all applications.

With this data generation scheme it is in principle possible to generate an infinite stream of data during training. However, we differed from this by generating a large training set, as it was more practical wrt. computational time when comparing different models.

For 1D to 3D the PDFs used for evaluating the presented approach are generated from real-world data. For this we used subsets of a stock market dataset in 1D,[1] of Imagenet in 2D [31] and of DeepLesion in 3D [36]. To accomplish that, the stock values over time and the gray-scale pixel/

---

[1] *Huge Stock Market Dataset* (2017), authored by Boris Marjanovic. Retrieved from https://www.kaggle.com/borismarjanovic/price-volume-data-for-all-us-stocks-etfs/version/3 (Nov. 2019).

voxel intensities are treated as discrete density functions. To gain continuous data from the otherwise discrete densities, the values at arbitrary positions are interpolated from the surrounding positions. For larger dimensions, the PDFs for evaluation were instead purely synthetic, as we are not aware of available datasets. The generation followed the same scheme as for the training data, but with a different set of base functions, constrained on certain characteristics.

The training sample sets for each dimensionality comprised 1000 samples, whereby every sample contained 1000 points for 1D and 5000 points for the rest. The validation set during training was a quarter of the respective training set, split before training. To evaluate the generalization capability, additional sets of synthetic samples were generated only including certain characteristics in the sampled functions or excluding them from the complete function set.

## 5 Evaluation

To evaluate the proposed DDE method, we use it to infer PDFs for synthetic and real-world data unseen during training, as well as single analytical PDFs in 1D. The performance of every prediction is quantified by the total computing time and two distance metrics, the mean pairwise squared error (MSE) and the Kullback–Leibler (KL) divergence [15, 22] as distance metrics between the true PDF values. Furthermore, we consider the $p$ value of the two sample Kolmogorov–Smirnov test, which is an estimate on the distribution itself. In the latter we compare the input distribution with a distribution sampled from the estimated PDF, where the $p$ value is an expression on how likely it is that the two distributions come from the same PDF. The DDE model is trained only once for a given domain dimensionality, with the trained state then directly applied to arbitrary sample distributions with same dimensionality. Thus, the time it takes to train the model is not regarded in the evaluation, because we are only interested in the time it takes to get an estimation for any given distribution. A dependence of the reported computing times with respect to varying implementation should only be of a minor order, as all tested methods are, at least for their time-wise most complex parts, run with CPU implementations from widely used and advanced software libraries. We did not engage in writing sophisticated GPU implementations for competing methods, as the main goal of this paper is to demonstrate that learned density estimation is highly accurate and can serve as an off the shelf tool for data scientists, even when it is only trained once.

To perform a meaningful evaluation, we compare our estimations to several frequently used density estimators available in Python and R. These are chosen, as Python is the most used programming language for the data-based sciences, directly followed by R,[2] which especially is the standard tool to solve statistical problems with many implementations unmatched by other languages. For Python we are comparing against a naive KDE implementation with Silverman's rule of thumb [34] for the bandwidth estimation (KDE), a Gaussian mixture model for density estimation (GMM) and a variational Bayesian Gaussian mixture model [4] (BGMM). For R we are comparing against the implementation of KDE with the plugin bandwidth estimator $R_{pi}$ [35], the smoothed cross-validation bandwidth estimator $R_{scv}$ [11, 20], the least squares cross-validation bandwidth estimator $R_{lscv}$ [5, 30], the normal mixture bandwidth $R_{nm}$ [37] and the normal scale bandwidth $R_{ns}$ [7], as well as the R implementation of Farmer and Jacob's PDF estimator [13] (FJE) in 1D, as the latter is only defined in 1D. In addition, we are comparing to a TensorFlow implementation of the masked autoregressive flow [26], which is a recent approach for deep density estimation. As not all methods can be presented in the plots, we have chosen to display only the estimators which showed in any test either the best result, or a good trade-off between different metrics, such as between time and MSE. The exclusion of this is $R_{scv}$, which is not represented, as it produces almost identical results to $R_{pi}$. Furthermore, we omitted $R_{hlscv}$ which, while producing good results in some cases, fails horribly for single distributions, making it impossible to compare in the chosen plots. All other results are reported in the tables in Appendix 3.

A summary of the test results for domain dimensionalities $d \in \{1, 3, 5, 10, 30\}$ is presented in Fig. 4 with the MSE and total computational time of estimation over the dataset and in Fig. 5 with the KL divergence, while the numerical values are presented in Appendix 3. For higher dimensions $\gtrsim 5C$ DDE shows the best MSE and is comparable with respect to the KL divergence, while being the fastest method in most cases. For smaller dimensions $\lesssim 5$ there is a give and take of time and accuracy between DDE and the competing methods. While in 3D DDE shows the worst KL divergence for smaller distribution sizes, it improves in accuracy for larger distribution sizes, which we could not equally observe in the other methods. The same relation is apparent for MSE, while the different scores are more comparable here. Regarding computing time, DDE is among the fastest method only beaten by vanilla KDE and also by $R_{ns}$ for large sample sizes. The former is significantly worse regarding the MSE, and while better for low distribution sizes regarding the KL

---

[2] *Visualization of Kaggles 2018 data science survey* (2018), authored by Kaggle and sudhirnl7 at https://www.kaggle.com/sudhirnl7/data-science-survey-2018 (retrieved October 2020).

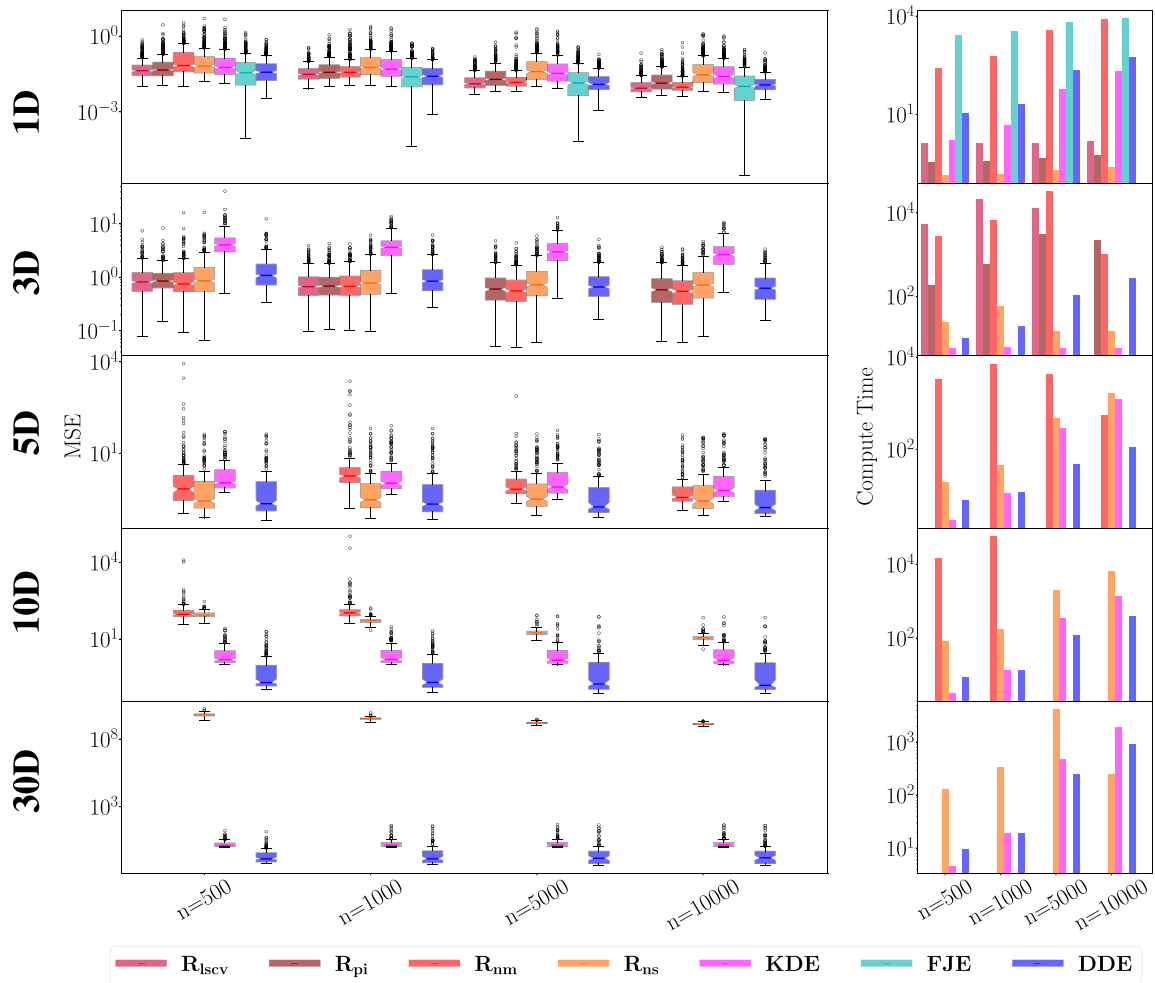

**Fig. 4** MSE per sampled distribution (*left*) and the total computing time (*right*) on a logarithmic scale—in both cases lower is better. The MSE plot shows the median as bar with a notched box showing the interquartile range (IQR) and the 1.5 IQR range on the whiskers with outliers as black circles. The methods shown are either the best methods with respect to accuracy and computing time or show the best trade-off between those two. Sometimes better scoring methods (such as $R_{nm}$) could not be tested for higher dimensions, as they simply take to much time. In 1D and 3D the tests were conducted on the stock market and DeepLesion test sets, respectively, and for higher dimensions on synthetically generated datasets

divergence, gets again worse for larger distribution sizes. The latter is better in both MSE and KL divergence, getting passed by DDE only for large sample sizes. In 1D we find that the best estimator is FJE which shows the best MSE score, and on average the best KL divergence for all tested distribution sizes, while it also contains some outliers with worse estimates regarding the latter. Unfortunately, FJE is consistently the slowest in all performed tests. In comparison with the other tested methods, DDE scores better regarding both the KL divergence and MSE, where the other methods become comparable only for larger distribution sizes. Regarding speed, DDE is slower than most methods in 1D, which is caused by the additional smoothing operation. Thus, for small dimensionalities DDE cannot for all tests be generally regarded as the best method on the evaluated distributions, but can neither be regarded as worse than the other evaluated techniques.

In the next step, we take a closer look on the estimations for distinct distributions in 1D. For each we present the estimate by $R_{nm}$, KDE, FJE and DDE, both with post applied smoothing and without, for distributions of size $n = 500$ and $n = 5000$ along with the MSE, KL divergence, $p$ value and estimation time. The distributions we are estimating are the five test distributions of Farmer and Jacobs [13].

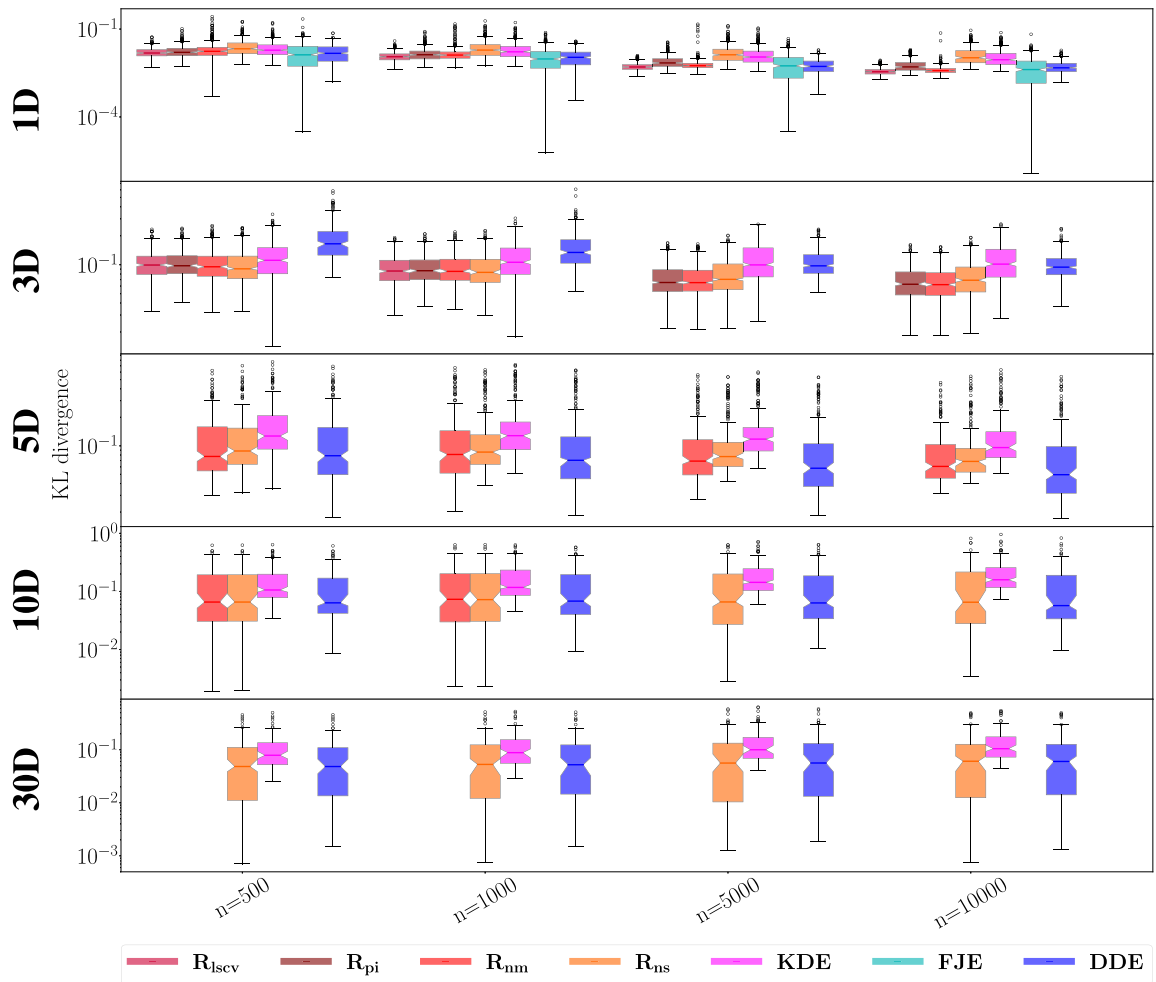

**Fig. 5** KL divergence per sampled distribution on a logarithmic scale—lower is better. The plot shows the median as bar with a notched box showing the interquartile range (IQR) and the 1.5 IQR range on the whiskers with outliers as black circles. In 1D and 3D the tests were conducted on the stock market and DeepLesion test sets, respectively, and for higher dimensions on synthetically generated datasets

## 5.1 Gamma distribution

The gamma distribution $p(x) = \frac{1}{\sqrt{\pi}x}e^{-x}$ presents the significant feature of a singularity at $x \to 0$, shown in Fig. 6. Going through the estimators $R_{nm}$ fails in estimating the PDF for both tested distribution sizes, which is however not distinctively apparent in the numeric scores, as it finds a good estimate for the actual divergence. KDE can fit the tail of the distribution, but fails for the divergence, which is also represented by the bad scores in all metrics. FJE, while being the slowest method, finds a good estimate for the divergence, but cannot fit the overall shape of the PDF well. While DDE can estimate distribution well for most parts, it fails to fit the divergence for small distribution sizes. The smoothing has only a minor effect on DDE for this distribution.

## 5.2 Sum of two Gaussians distribution

The sum of two Gaussians distribution $p(x) = \frac{7}{10}\mathcal{N}(x|\mu = 5, \sigma = 3) + \frac{3}{10}\mathcal{N}(x|\mu = 0, \sigma = \frac{1}{2})$, where $\mathcal{N}$ denotes the Gaussian distribution with mean $\mu$ and standard deviation $\sigma$ is a standard multimodal distribution with soft tails, shown in Fig. 7. For this case, $R_{nm}$ finds a good estimate, which follows the general structure of the PDF, with some high-frequency errors. This is also apparent by the significantly small MSE and KL divergence and the large $p$ value. While KDE can also reproduce the general structure of the PDF, both peaks are under-estimated, causing significantly lower scores. The quality of the FJE estimate is both visually and numerically in between those of KDE and $R_{nm}$. While ii recovers the general structure of the PDF also for $n = 500$, the sharp peak is still under-estimated, while the left bump is shifted and narrower than it should be. For $n = 5000$, the sharp peak is estimated well

**Fig. 6** Results on the test of the gamma distribution in 1D. On top a plot of the true PDF and the estimates of $R_{nm}$, KDE, FJE and DDE with and without post applied smoothing. For the plot on the left the estimators learned the distribution of a sample size of $n = 500$ and $n = 5000$ on the right. Below are the tabulated metrics of MSE, KL divergence, $p$ value and the computational time

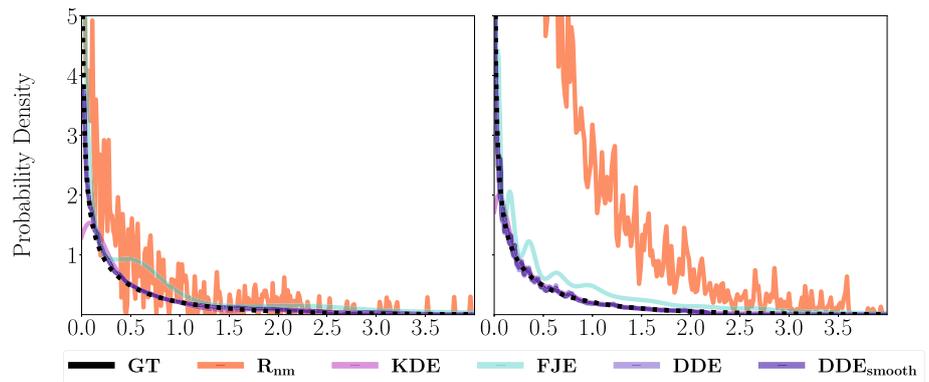

**n = 500**

| Method | MSE | KL div. | p-value | time |
|---|---|---|---|---|
| $\mathbf{R_{nm}}$ | 1.893e+04 | 5.669e-01 | 8.167e-05 | 4.354e-01 |
| **KDE** | 2.114e+04 | 9.071e-01 | 4.212e-04 | 8.423e-03 |
| **FJE** | 1.895e+04 | 5.178e-01 | 2.755e-06 | 2.679e+01 |
| **DDE** | 2.045e+04 | 6.445e-01 | 5.869e-02 | 2.535e-01 |
| $\mathbf{DDE_{smooth}}$ | 2.045e+04 | 6.445e-01 | 4.950e-02 | 2.582e-01 |

**n = 5,000**

| Method | MSE | KL div. | p-value | time |
|---|---|---|---|---|
| $\mathbf{R_{nm}}$ | 1.178e+05 | 6.128e-01 | 3.403e-08 | 1.657e+00 |
| **KDE** | 9.289e+04 | 8.876e-01 | 1.023e-32 | 7.181e-01 |
| **FJE** | 8.381e+04 | 3.006e-01 | 4.196e-29 | 4.324e+00 |
| **DDE** | 8.712e+04 | 3.131e-01 | 5.385e-03 | 5.637e-01 |
| $\mathbf{DDE_{smooth}}$ | 8.712e+04 | 3.131e-01 | 6.174e-03 | 6.672e-01 |

**Fig. 7** Results on the test of the sum of two Gaussians distribution in 1D. On top a plot of the true PDF and the estimates of $R_{nm}$, KDE, FJE and DDE with and without post applied smoothing. For the plot on the left the estimators learned the distribution of a sample size of $n = 500$ and $n = 5000$ on the right. Below are the tabulated metrics of MSE, KL divergence, $p$ value and the computational time

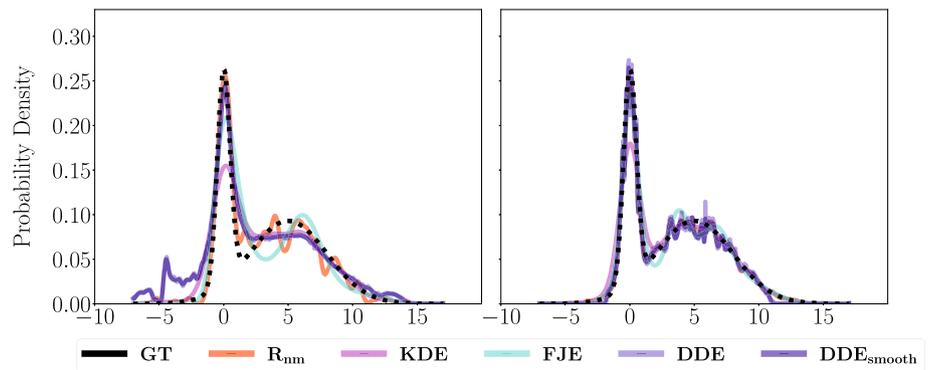

**n = 500**

| Method | MSE | KL div. | p-value | time |
|---|---|---|---|---|
| $\mathbf{R_{nm}}$ | 9.961e-02 | 8.835e-03 | 6.659e-01 | 1.355e-01 |
| **KDE** | 1.206e+00 | 5.238e-02 | 1.294e-01 | 8.658e-03 |
| **FJE** | 4.323e+00 | 2.931e-01 | 3.294e-01 | 2.858e+01 |
| **DDE** | 1.984e-01 | 2.004e-02 | 8.167e-05 | 3.128e-01 |
| $\mathbf{DDE_{smooth}}$ | 2.188e-01 | 2.209e-02 | 4.212e-04 | 3.276e-02 |

**n = 5,000**

| Method | MSE | KL div. | p-value | time |
|---|---|---|---|---|
| $\mathbf{R_{nm}}$ | 4.426e-02 | 2.379e-03 | 7.764e-01 | 9.668e-01 |
| **KDE** | 5.776e-01 | 2.193e-02 | 4.076e-03 | 7.759e-01 |
| **FJE** | 9.927e-02 | 6.972e-03 | 6.104e-01 | 1.323e+00 |
| **DDE** | 8.590e-02 | 6.317e-03 | 3.928e-01 | 3.276e-01 |
| $\mathbf{DDE_{smooth}}$ | 8.088e-02 | 5.457e-03 | 8.356e-02 | 3.601e-01 |

and the left bump is roughly estimated, while the shape here is more akin to actual features, as the bump is split in two, while the errors of $R_{nm}$ are more akin to noise. While DDE can estimate the structure of the PDF correctly, it estimates a wrong feature to the left of the strong peak for $n = 500$ and the tails vanish to quickly for both tested distribution sizes. Hence, the scores on MSE and KL divergence still show both, good results for both $n$, while the p value for $n = 500$ is quite low. Again the smoothing has only a minor, while visible and measurable effect.

### 5.3 Five fingers distribution

The five fingers distribution $p(x) = w \sum_{k=1}^{5} \frac{1}{5} \mathcal{N}(x | \mu = \frac{2k-1}{10}, \sigma = \frac{1}{100}) + (1 - w)$ with $w = 0.5$ contains five sharp Gaussian peaks, shown in Fig. 8. This type of distribution is a good test for estimators, as the sharp peaks and only local expression of the PDF with vanishing probability on large ranges of the domain, pose a difficult challenge for density estimation. In particular $R_{nm}$ fails again for the estimation of this PDF, producing an almost flat line for both distribution sizes. KDE scores similarly poorly, while at least estimating the Gaussian's to some degree, albeit far to under-expressed. For this distribution FJE does not manage to recover the five peaks, but instead estimates only three, leading to a large KL divergence and a very low

$p$ value. We note that this faulty estimate is maybe caused by a faulty estimation in the R package, as the estimate of this distribution in the original paper of Farmer and Jacobs is better. The estimation of DDE for $n = 500$ is not close to a PDF in this example, as the area under the curve is by far too large, but it is the only method able to reproduce the general structure of the PDF with sharp peaks, while the virtual peaks between the actual Gaussians are a wrong feature estimated for small sample sizes, caused by the roughly symmetric distribution of sampled points around them. For $n = 5000$ the shape of the PDF is better estimated, with only the peaks being too high, causing the lowest KL divergence and highest p value. Here the effect of smoothing is again very small.

### 5.4 Cauchy distribution

The Cauchy distribution $p(x) = \frac{b}{\pi(x^2 + b^2)}$ has heavy tails. The extreme statistics of the Cauchy distribution are generally a difficult problem for density estimation, shown in Fig. 9. Here $R_{nm}$ and KDE give visually almost identical estimates with good scores, while for $n = 5000$ the former shows better distance scores, while the latter has a significantly higher $p$ value. Again, the estimate of FJE is far away from the true PDF and by far the worst of the compared estimates, while again, this distribution was better

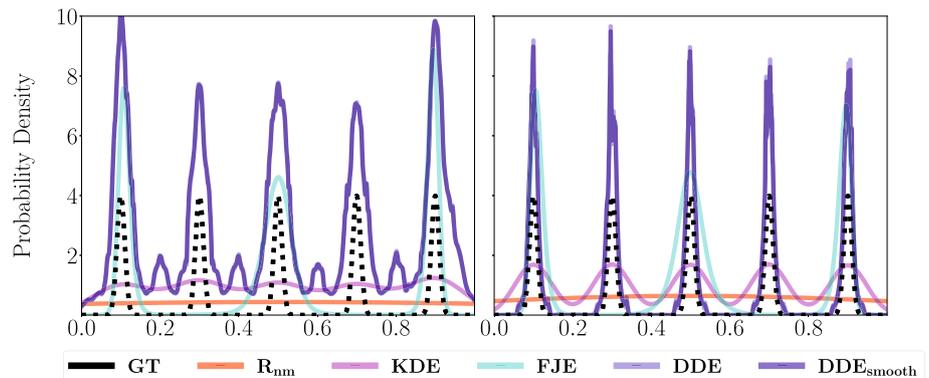

**Fig. 8** Results on the test of the five fingers distribution in 1D. On top a plot of the true PDF and the estimates of $R_{nm}$, KDE, FJE and DDE with and without post applied smoothing. For the plot on the left the estimators learned the distribution of a sample size of $n = 500$ and $n = 5000$ on the right. Below are the tabulated metrics of MSE, KL divergence, $p$ value and the computational time

**n = 500**

| Method | MSE | KL div. | p-value | time |
|---|---|---|---|---|
| $R_{nm}$ | 7.188e+00 | 1.490e-01 | 1.159e-03 | 2.215e-01 |
| KDE | 4.258e+00 | 1.420e-01 | 2.990e-03 | 8.642e-03 |
| FJE | 6.581e+00 | 9.767e-01 | 5.832e-03 | 1.305e+00 |
| DDE | 2.313e+01 | 7.563e-02 | 4.159e-02 | 3.114e-02 |
| $DDE_{smooth}$ | 2.311e+01 | 7.549e-02 | 7.228e-03 | 3.348e-02 |

**n = 5,000**

| Method | MSE | KL div. | p-value | time |
|---|---|---|---|---|
| $R_{nm}$ | 6.216e+00 | 1.572e-01 | 2.065e-23 | 3.584e-01 |
| KDE | 2.496e+00 | 1.328e-01 | 1.192e-12 | 7.186e-01 |
| FJE | 8.113e+00 | 5.877e-01 | 4.865e-27 | 1.817e+00 |
| DDE | 8.988e+00 | 6.245e-03 | 5.224e-02 | 3.206e-01 |
| $DDE_{smooth}$ | 8.984e+00 | 6.181e-03 | 5.767e-03 | 3.683e-01 |

**Fig. 9** Results on the test of the Cauchy distribution in 1D. On top a plot of the true PDF and the estimates of $R_{nm}$, KDE, FJE and DDE with and without post applied smoothing. For the plot on the left the estimators learned the distribution of a sample size of $n = 500$ and $n = 5000$ on the right. Below are the tabulated metrics of MSE, KL divergence, $p$ value and the computational time

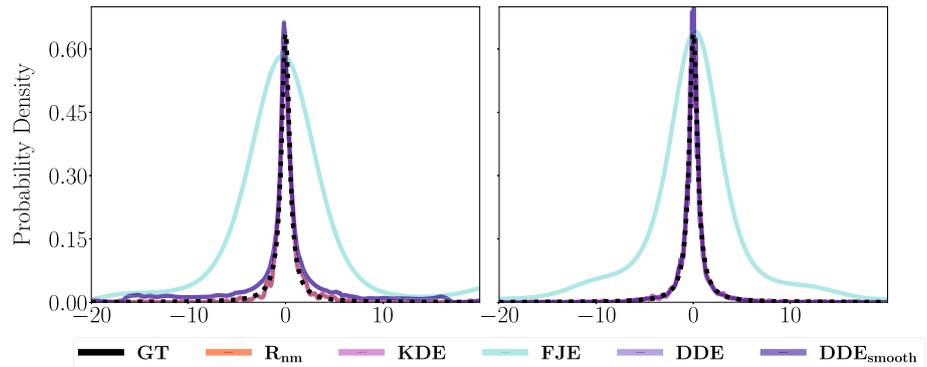

**n = 500**

| Method | MSE | KL div. | p-value | time |
|---|---|---|---|---|
| **$R_{nm}$** | 9.539e+00 | 6.020e-03 | 9.351e-01 | 1.069e-01 |
| **KDE** | 9.676e+00 | 6.037e-03 | 9.351e-01 | 7.955e-03 |
| **FJE** | 4.588e+02 | 2.866e-01 | 1.280e-21 | 3.607e+01 |
| **DDE** | 1.635e+01 | 1.292e-02 | 1.975e-06 | 3.156e-02 |
| **DDE$_{smooth}$** | 1.633e+01 | 1.291e-02 | 6.125e-05 | 3.263e-02 |

**n = 5,000**

| Method | MSE | KL div. | p-value | time |
|---|---|---|---|---|
| **$R_{nm}$** | 1.655e+00 | 1.163e-03 | 3.275e-01 | 7.198e-01 |
| **KDE** | 1.019e+00 | 8.665e-04 | 3.657e-01 | 6.281e-01 |
| **FJE** | 5.271e+02 | 2.500e-01 | 7.916e-150 | 1.469e+00 |
| **DDE** | 9.918e+00 | 5.380e-03 | 1.420e-01 | 3.262e-01 |
| **DDE$_{smooth}$** | 9.914e+00 | 5.352e-03 | 1.627e-01 | 3.520e-01 |

estimated by Farmer and Jacobs. DDE over estimates the tails of the distribution for $n = 500$ with a slightly to high peak in the center. While this also causes bad scores, the estimation becomes much better for larger sample sizes. In this case the effect of smoothing is visually not noticeable for both sample sizes and even causes worse metrics for $n = 5000$.

### 5.5 Discontinuous distribution

The discontinuous distribution

$$p(x) = \begin{cases} \frac{4}{5}, & \text{if } x < 0.3 \quad \text{or } x > 0.8 \\ 1, & \text{if } 0.4 < x < 0.5 \\ \frac{5}{4}, & \text{else} \end{cases}$$

defined on the range [0, 1], poses the problem of discontinuities in the PDF with heavy edges, as shown in Fig. 10. While no method can estimate the sharp edges of this distribution well for $n = 500$, almost all methods can estimate the larger bump of the distribution, with only FJE producing one smooth curve offers the entire PDF. DDE and KDE are closer to the valley of the distribution, but $R_{nm}$ can also estimate the second bump as a noticeable feature. For $n = 5000$ KDE still produces an overly smooth estimate, while the unsmoothed result of DDE contains many noisy spikes. Here, the smoothing has the strongest effect, leading to an estimate which is visually similar to that of $R_{nm}$, while showing the best distance scores. FJE does now estimate the larger bump of the distribution, while showing no other features of the true PDF, apart from reproducing the heavy edges on $x = 0$ and $x = 1$ better than KDE, as it does not quickly degrade toward the end of the distributions.

To summarize these five results, we can say that DDE has difficulties both in estimating sharp peaks as well as long tails, where in the former it produces to high estimates and in the latter vanishes to quickly. Still, DDE is the only method which reproduced the functional shape of the target PDF in all tests, which makes it a good method for estimating such.

### 5.6 Examples for higher domain dimensions

To compare the actual shape of the density estimates in higher dimensions, we present examples of estimations in 2D and 3D domains in Fig. 11. As a depiction of the estimations becomes sub-optimal already for 3D, where we plot a slice of the volume projected onto the $xy$-plane, we cannot show them for higher dimensions. In the figure we present the results of $R_{pi}$, $R_{ns}$, $R_{nm}$ and DDE, both with their respective estimate, as well as an error image between the true PDF and the estimate. Both in 2D and 3D, $R_{pi}$ and $R_{ns}$ produce very smooth estimates. While this may be beneficial on areas with a rather flat distribution, it becomes certainly a problem in regions with dominant features, which cannot be estimated well. DDE and $R_{nm}$ on the other hand give closer estimates to the PDF. Especially in 2D the estimates are similar, while $R_{nm}$ appears as a smoother version of DDE, which however seems to be an unwanted feature both regarding the visual quality, as well as the MSE, although only marginally. This is even more so true for 3D, where only DDE manages to estimate the sharp features of the distribution, which becomes also visible in the error image, where all KDE-based estimators still show more distinct structures, while it is more distributed for DDE and akin to noise. Note that these error images are normalized on their own range, not with respect to a common factor and should thus be compared along the actual error value. Thus, although comparable to $R_{nm}$ in the total accuracy, DDE appears to produce an estimate which is visually closer to the real PDF. This is essentially the same result as in 1D, where the metrics were on par or sometimes worse than other estimators, but still the estimate was visually more akin to the PDF.

**Fig. 10** Results on the test of the discontinuous distribution in 1D. On top a plot of the true PDF and the estimates of $R_{nm}$, KDE, FJE and DDE with and without post applied smoothing. For the plot on the left the estimators learned the distribution of a sample size of $n = 500$ and $n = 5000$ on the right. Below are the tabulated metrics of MSE, KL divergence, $p$ value and the computational time

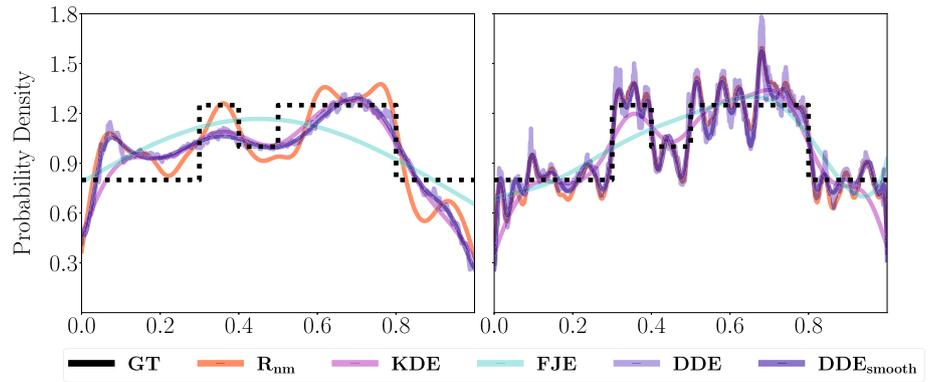

| n = 500 | | | | |
|---|---|---|---|---|
| Method | MSE | KL div. | p-value | time |
| $R_{nm}$ | 2.306e-02 | 1.240e-02 | 6.126e-01 | 4.190e-01 |
| KDE | 2.171e-02 | 1.181e-02 | 5.600e-01 | 8.724e-03 |
| FJE | 2.762e-02 | 1.260e-02 | 5.600e-01 | 3.536e+00 |
| DDE | 2.708e-02 | 1.513e-02 | 5.600e-01 | 3.108e-02 |
| $DDE_{smooth}$ | 3.062e-02 | 1.694e-02 | 8.192e-01 | 3.323e-02 |

| n = 5,000 | | | | |
|---|---|---|---|---|
| Method | MSE | KL div. | p-value | time |
| $R_{nm}$ | 1.157e-02 | 5.122e-03 | 9.125e-01 | 5.238e+00 |
| KDE | 1.202e-02 | 6.813e-03 | 5.770e-01 | 7.724e-01 |
| FJE | 1.246e-02 | 5.800e-03 | 3.657e-01 | 1.659e+01 |
| DDE | 1.511e-02 | 6.403e-03 | 8.223e-01 | 3.189e-01 |
| $DDE_{smooth}$ | 1.142e-02 | 4.866e-03 | 2.700e-01 | 3.417e-01 |

**Fig. 11** Examples of the best tested density estimators $R_{pi}$, $R_{nm}$, $R_{ns}$ and DDE for distributions with 10,000 points each. Examples for 2D (*top*) and 3D (*bottom*). For 2D and 3D the ground truth PDF is given on the left. Below each estimation the respective error image is given with MSE values on top

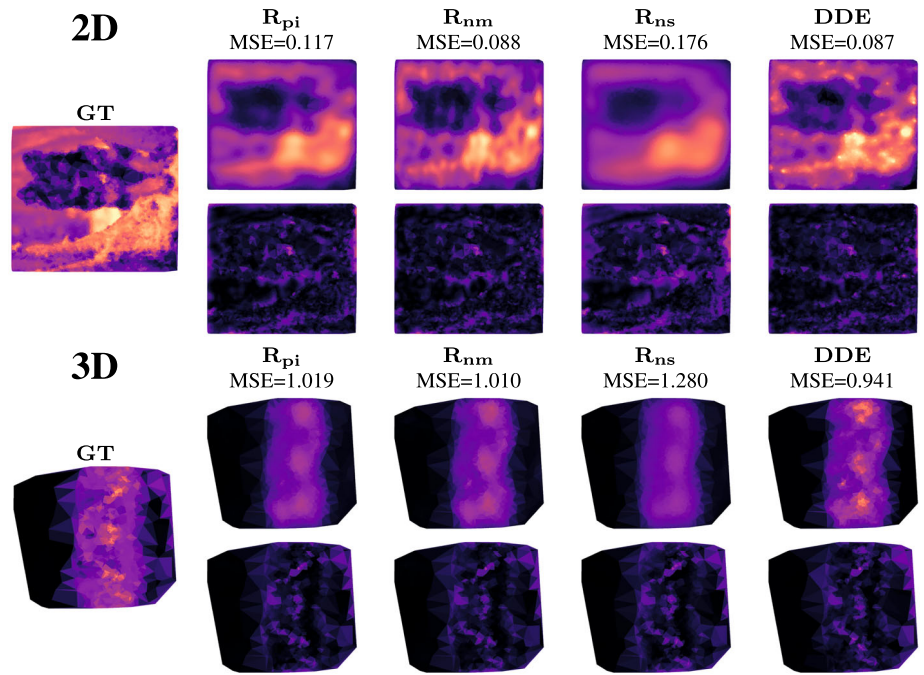

### 5.7 Generalizability evaluation

In this section, we describe our investigations toward the generalization of our proposed models to other data. Therefore, we compared models trained on subsets of the real-world data to the same models trained on synthetic data. During this comparison, we tested all trained models on the real-world test sets as well as different synthetic test sets as visualized in Fig. 12. We have further varied the distribution sizes, whereby all test distributions are disjoint from the training distributions. Note that the PDFs generated from real-world data are expected to be similar among train and test sets with regard to the functional shape. The results of our investigations show that the models trained on synthetic data perform equally or better than the models trained on real-world datasets, except for 3D where the real-world models always score better. We attribute this difference for the 3D case to the specific characteristics of the DeepLesion dataset. The dominant silhouette edges, which occur in the entire dataset, are hard to fit, but inherently learned by the, respectively, trained model, for exactly these type of edges at those positions, thus overfitting it to this kind of data. On the synthetic test distributions however, both types of model score similarly in 2D and 3D, while in 1D again the models trained on synthetic data score significantly better. This indicates that the models show a good generalization capability while also having space for additional specialization. The models trained on real-world data in 3D show comparable results on the synthetic test sets, but perform systematically better on the real-world test sets. We ascribe the fact that the models trained on real-world data perform worse on synthetic data in 1D, to a lacking diversity in the data of the real-world distributions in 1D. As this systematically lower score is also apparent on the real-world data, it could additionally indicate that the models' capacity was large enough to not only overfit similar functional shapes, but also the specific PDFs. This is possible in that case, as the model does not need to learn information about global shapes, before it can address local features, but instead learns only the local features based on the common global shape of the entire dataset.

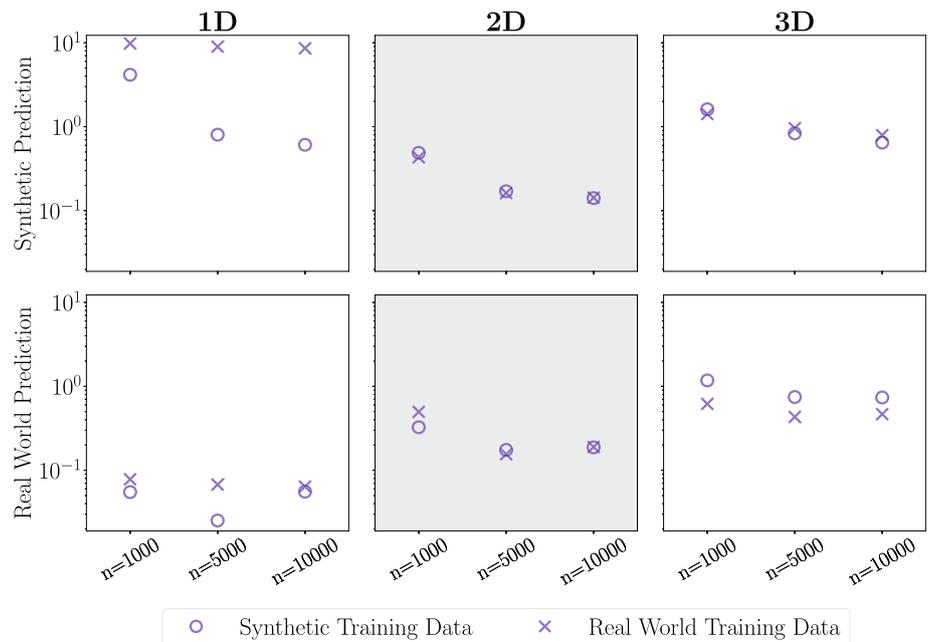

**Fig. 12** Comparison of the MSE score (on a logarithmic scale, lower means better) of models trained on synthetic data (*circle*) and such trained on subsets of the real-world dataset (*cross*), evaluated on synthetic (*top*) and real-world data (*bottom*). All test datasets are disjoint to the respective training datasets, but stem from the same respective corpus of data. Every tick represents the mean MSE over a complete dataset for one $n \in \{1000, 5000, 10,000\}$, where $n$ is the number of points per sample

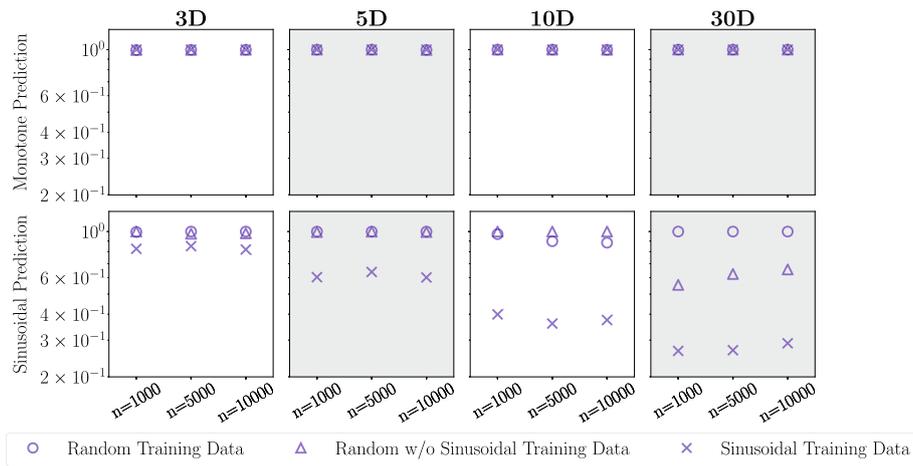

**Fig. 13** Comparison of the MSE score (on a logarithmic scale, lower means better) of models trained on synthetic data with random (circle), sinusoidal (cross) and random without sinusoidal characteristics (triangle), evaluated on synthetic data with monotone (*top*) and sinusoidal characteristics (*bottom*). All test datasets are disjoint to the respective training datasets but stem from the same respective corpus of data. Every tick represents the mean MSE over a complete dataset for one $n \in \{1000, 5000, 10,000\}$, where $n$ is the number of points per sample

For an additional test, we trained models on datasets comprising functions with random, sinusoidal and random with the exclusion of sinusoidal characteristics, with results in Fig. 13. In this test we can see that all models score roughly similar on the monotone dataset, indicating that this specific functional shape can also be well estimated by models trained exclusively on other functional shapes. As expected, the sinusoidal trained models score always significantly better on the sinusoidal datasets. Along this it is important to note that the model trained without sinusoidal functions scores worse on the sinusoidal PDFs in 10D than the one trained on random functions, but significantly better in 30D. This indicates that the specific functional shape must not be apparent in the training data in form of the 1D base functions, but that the important local characteristics can be generated by the random combination of the base functions. Nevertheless, this evaluation also shows that the generation of training data in high dimensions shows great potential for advancement, as the difference of the models trained on sinusoidal data to all other models gets larger with higher dimensions.

## 5.8 Complete data space estimation

Any method used for the process of estimating PDFs of arbitrary distributions has to estimate not only the areas of high probability accurately, but the complete data space. The difficult task in this is the correct estimation of areas with a very faint or zero probability. Examples for this task are presented in Fig. 14 for the same methods as before. For the compared methods we can see that DDE and $R_{\mathrm{nm}}$ can extrapolate to the upsampled regions much better than the other methods, caused by the overly smooth estimation of the complete data space by those. Even for the rather large sample sizes shown, $R_{\mathrm{ns}}$ predicts much too smooth estimations, missing every feature of the true PDF. Also $R_{\mathrm{pi}}$ seems to be unable to correctly estimate the parts of the PDF with a strong gradient, while it is able to correctly estimate the larger regions of zero probability in the bottom example. The estimations of $R_{\mathrm{nm}}$ and DDE can hardly be differentiated. Although $R_{\mathrm{nm}}$ shows a slightly better score, the visual quality of the estimations appears equivalent, with the only difference that $R_{\mathrm{nm}}$ estimates a slightly smoother PDF which can be preferred or not, while DDE again holds the benefit of being much faster than $R_{\mathrm{nm}}$. Also FJE cannot correctly estimate the periodic function on the top, where the valleys and peaks are shifted in some cases. While the estimate produced by FJE is still better than of vanilla KDE or $R_{\mathrm{ns}}$, it is not as accurate as DDE, $R_{\mathrm{pi}}$ or $R_{\mathrm{nm}}$. In the lower case its estimate is close to the last mentioned methods, but still shows some distinct variations from these and the true PDF.

## 5.9 Local shape dependence

We also evaluated the dependency of the different estimators on the local density around a given query point. For this, we constructed a range of PDFs shown in Fig. 15, for which the position $t$ such that $p(t) = 1$ is known. The definition of the respective functions from left to right and top to bottom is given in Table 1. Given distributions with sizes $n = 500$ and $n = 10,000$, every estimator was evaluated on $t$. We again present the values of the best estimators in Table 1. We show the estimate at the query point

**Fig. 14** Examples of the estimated PDF values for the complete data space in 1D for $R_{pi}$, $R_{nm}$, $R_{ns}$, KDE, FJE and DDE alongside the true PDF values for the complete data space. The distributions were sampled with $n = 5000$ points each from synthetic PDFs disjoint to the synthetic training data

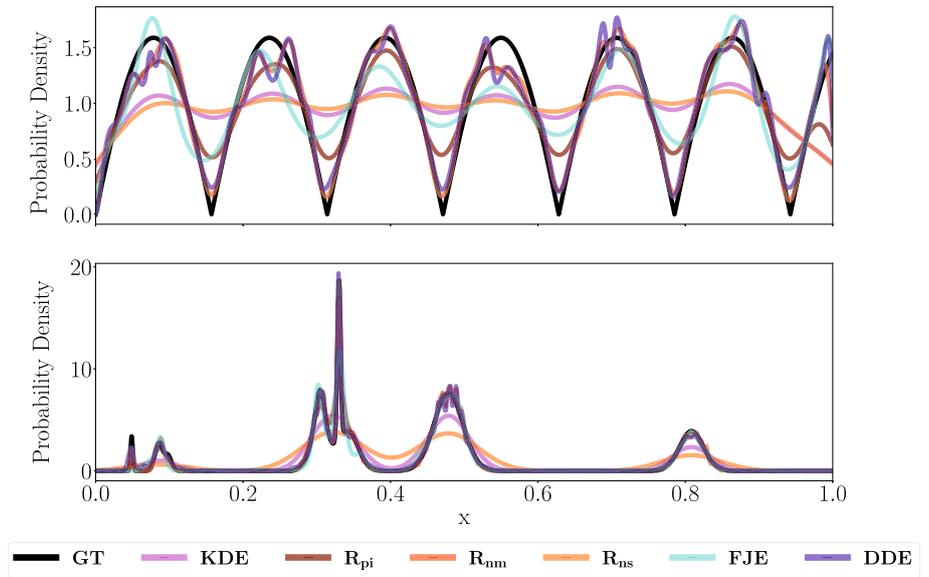

**Fig. 15** A set of 9 PDFs with 1D domain. For every PDF we have a distribution which is used get the estimate $p(t)$ at position $t$, with the ground truth values of $p_{GT}(t) = 1$

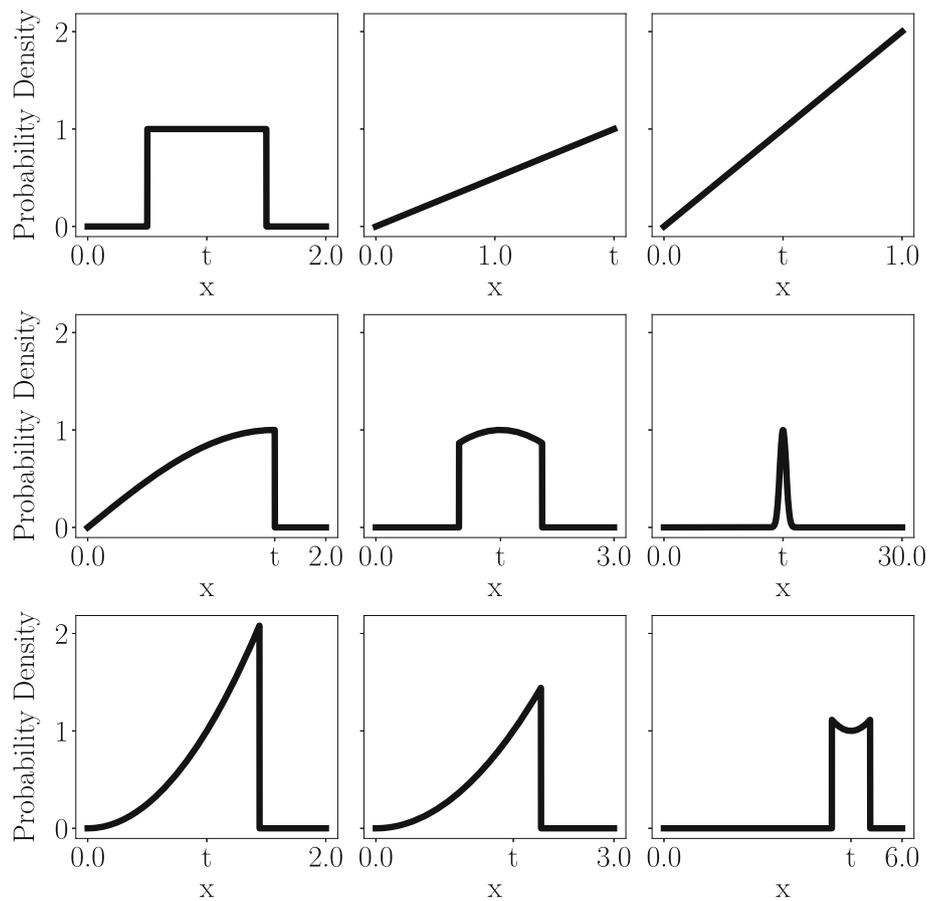

**Table 1** Numerical results of the tests for dependence on local shape of the distribution's underlying PDF

| PDF $p(x)$ | $R_{pi}$ | $R_{nm}$ | $R_{ns}$ | KDE | FJE | DDE |
|---|---|---|---|---|---|---|
| $n = 500$ | | | | | | |
| $1$   if   $0.5 < x < 1.5$ | 0.83 | 0.75 | 0.87 | 0.80 | **0.99** | 0.88 |
| $\frac{x}{2}$   if   $x < 2$ | 0.45 | **0.56** | 0.45 | 0.41 | 0.25 | 0.46 |
| $2x$   if   $x < 1$ | 0.98 | 0.88 | **1.00** | 0.59 | 1.17 | 1.01 |
| $\sin x$   if   $x < \frac{\pi}{2}$ | 0.49 | 0.42 | 0.50 | 0.47 | **0.88** | 0.49 |
| $\sin x$   if   $\frac{\pi}{3} < x < \frac{2\pi}{3}$ | 1.01 | **1.00** | 1.01 | 0.90 | 0.94 | 0.97 |
| $e^{\frac{-(x-\mu)^2}{2\sigma^2}}$   if   $x < 30$ | 0.95 | 0.94 | 0.94 | 1.02 | 0.93 | **1.00** |
| $x^2$   if   $x < 3^{\frac{1}{3}}$ | 0.79 | **0.96** | 0.86 | 0.92 | 0.62 | 0.85 |
| $\frac{x^2}{3}$   if   $x < 9^{\frac{1}{3}}$ | 0.92 | 0.86 | **0.96** | 0.82 | 0.49 | 0.91 |
| $\sin x + 2$   if   $\frac{3\pi}{2} - \frac{\pi}{\alpha} < x < \frac{3\pi}{2} + \frac{\pi}{\alpha}$ | **1.01** | 0.44 | **0.99** | 1.22 | 1.03 | 0.96 |
| Mean | 0.82 | 0.76 | **0.84** | 0.79 | 0.81 | **0.84** |
| SD | **0.20** | 0.21 | 0.21 | 0.25 | 0.28 | **0.20** |
| $n = 10,000$ | | | | | | |
| $1$   if   $0.5 < x < 1.5$ | 1.01 | 1.11 | 0.99 | 1.28 | **1.00** | 1.02 |
| $\frac{x}{2}$   if   $x < 2$ | 0.50 | 0.52 | 0.49 | 0.36 | **1.06** | 0.50 |
| $2x$   if   $x < 1$ | 1.05 | 1.05 | 1.03 | 1.68 | 0.99 | **1.00** |
| $\sin x$   if   $x < \frac{\pi}{2}$ | 0.49 | 0.49 | 0.51 | 0.18 | **0.92** | 0.50 |
| $\sin x$   if   $\frac{\pi}{3} < x < \frac{2\pi}{3}$ | 0.98 | **1.02** | 0.96 | 0.51 | 0.95 | 0.94 |
| $e^{\frac{-(x-\mu)^2}{2\sigma^2}}$   if   $x < 30$ | **0.98** | **0.98** | **0.98** | 1.05 | 0.92 | 1.06 |
| $x^2$   if   $x < 3^{\frac{1}{3}}$ | 0.97 | 0.85 | **1.01** | 1.11 | 0.54 | 0.81 |
| $\frac{x^2}{3}$   if   $x < 9^{\frac{1}{3}}$ | **1.01** | 0.93 | **0.99** | 1.35 | 0.47 | 0.87 |
| $\sin x + 2$   if   $\frac{3\pi}{2} - \frac{\pi}{\alpha} < x < \frac{3\pi}{2} + \frac{\pi}{\alpha}$ | 1.05 | 1.09 | **1.04** | 0.71 | **1.04** | 1.05 |
| Mean | 0.89 | 0.89 | 0.89 | **0.91** | 0.88 | 0.86 |
| SD | 0.21 | 0.21 | 0.21 | 0.47 | 0.28 | 0.21 |

All values are the respective density estimate at point $t$ with ground truth $p(t) = 1$. The mean and standard deviation of the Gaussian in row 6 are defined as $\sigma = \frac{1}{\sqrt{2\pi}}$, $\mu = 15$ and the value $\alpha$ in row 9 is defined as $\alpha = 6.52326761054738$. For every PDF $p(x) = 0$ for $x$ outside the defined ranges. The table on the top shows the results for distributions with size $n = 500$ and the bottom one for $n = 10,000$

$t$ for every estimator and the respective mean value and standard deviations. The estimate as well as the mean value should be close to 1, while the standard deviation should be as low as possible for an estimator who is robust with respect of the local shape. While DDE knows per definition only the information carried by the $k = 128$ closest points, the other estimators were still fed with the full distribution. As can be seen from the results in row 2 and 4, DDE struggles with correct estimates on heavy edges. The only estimator, out of the ones we compared here, solving this task is FJE, although not in all cases. Additionally, FJE produces significantly bad estimates for query points on a rising slope, visible in rows 7 and 8, which no other estimator does. Other than that, DDE shows results comparable to the other estimators. On average, DDE shows the best results for low sample sizes, but shows worse results for high sample sizes by the same margin. While we cannot deduce any significant advantage or disadvantage of DDE given the results of this analysis, it shows that DDE is on average at least just as resilient to the local shape regarding the estimation accuracy, as other estimators are.

## 6 Discussion

The problem of automated density estimation is crucial for many data analysis tasks and beyond. We presented a well-generalizing novel data-driven neural network approach to solve this problem. To achieve this task we have proposed the model architecture of DDE and a pipeline for the generation of synthetic PDF datasets, making generalizable training possible. All material for retraining, generating data and the actual implementation of DDE is publicly available.[3] With the provided implementation, every task can be completed by a single function call and is thus

---

[3]   https://github.com/trikpachu/DDE.

resulting in no barrier for potential users or researcher benchmarking their own results.

The comparison with the state-of-the-art density estimators in Sect. 5 indicates that first of all DDE shows the best general applicability in high dimensions, concerning that other methods are significantly slower, with on average better or comparable scores both in accuracy and speed with respect to the compared methods and evaluated datasets. While this is achieved by employing generalized models, even better results could be expected when retraining our estimator for the case at hand, in a test scenario where the same family of distributions has to be evaluated many times. This is visible from the results in Sect. 5.7, where we could achieve better scores for a model trained and evaluated on such data with strong global similarities. As we could show especially for 1D problems, our method does not predict the best estimates in all cases, but nevertheless always predicted estimates, reminiscent of the real PDF in structure, without failing completely for single distributions.

## 7 Conclusion

The DDE method poses a new, generally applicable tool for data-driven density estimation. While it produces good results, structurally reminiscent of the real PDF, it does not always show the best scores, while also the speed of the method could be enhanced. Thus we discuss here a few possible paths to further advance DDE. On the one hand, we have seen in Fig. 8 that albeit DDE produces visually good estimates of the PDF, in regard of the shape of it, it can be far away of a real PDF, as it is not normalized to one. While this happens only for very few cases, a normalization, preferably not as an expensive postprocessing step, should be employed somehow in the algorithm. Currently we employed only the MSE for our training objective. This could be enhanced by employing different metrics, perhaps more suited to the application toward probabilities, such as the KL divergence. Another example is the modification of the method to better respond to anisotropically distributed samples by respecting the edges of the distribution. In addition, the single model could be transformed into an ensemble of models, where first one model decides for the value of $k$ and then assigns the corresponding model, which would be similar to the current model formulation. This also allows for the estimation of distributions smaller than the default $k$. Additionally, two advancements of the synthetic function generation should be made. First, the set of base functions should be adapted to include more functional characteristics and a theoretically more sound set of complementary functions. Second, the generation of high dimensional synthetic data should be adapted, since the analysis of Sect. 5.7 has shown that the models generalization capability shrinks for higher dimensions. This could further increase the difference of accuracy in high dimensions with respect to the other estimators.

## Appendix 1: Hardware used for the evaluation

The networks are built and trained with TensorFlow 2.1, and all tests were conducted using an NVIDIA GTX 1080 GPU and an Intel Core i7-8700K CPU.

## Appendix 2: Set of base functions for the synthetic function generator

As described in Sect. 4 the generation of synthetic PDFs, from which to draw probability distributions with ground truth, takes a set of 1D base functions as building blocks. The functions from which those are chosen are listed in Table 2. Most of the functions take the parameters $R$ and $S$ as Input, where $R$ is a uniform random number in the range [0,1] $R \in U([1,0])$ and $S$ is the upper bound of the domain in the respective dimension. Here we explain how the selected functions satisfy the addressed necessary features of base functions, which are periodic or aperiodic features, (non-)-monotonicity, different degrees of slopes, signal peaks of varying degree, discontinuities, valleys, (non-)heavy edges, (non-)heavy tails, and semi-diverging features. While most proposed functions are aperiodic, the sine and cosine functions bear the feature of periodicity. Combined with other functions and expressed on varying domain extents, this results in different shapes of such periodicity. Monotonicity is expressed by the different linear or power functions. While the appearance of negative along positive monotonicity on its own is redundant for our model, it becomes important for the random combination with other asymmetric functions; thus, both are included. While basically every function produces different degrees of slopes, this is formally also encoded as well by $f(x) = x$, $f(x) = x^{2R}$ and $f(x) = x^2$, as the multiplicative combination of these leads to arbitrary degrees. The signal peaks are encoded by the different randomized Gaussians. Discontinuities in $f(x)$ are produced by the step functions, while discontinuities in $f'(x)$ are encoded by the norm of the sinusoidal functions, as well as by the maximum and minimum operators taking $x$ as argument. The latter also prohibit exploding values, while semi-diverging features are still expressed by the inverse functions of $x$ and $f(x) = x^2$ for large domain sizes. The functional valleys are again encoded by the sinusoidal functions. The heavy or non-heavy edges are encoded by basically every function. Most prominently do the step functions encode heavy edges, while all monotone functions have by default one heavy and one soft edge.

**Table 2** 1D base functions for the generation of synthetic PDFs

| Definition of function $f(x)$ | Parameters |
|---|---|
| $\frac{1}{1+e^{-Rx}}$ | |
| $\frac{2R}{\sqrt{2\pi\sigma^2}}\, e^{\frac{-(x-\mu)^2}{2\sigma^2}}$ | $(\mu, \sigma) \in \{(0.75RS,$ $\quad(0.5RS,$ $\quad(0.25RS,$ $\quad(0.75RS,$ $\quad(RS,$ $\quad(0.5RS,$ $\quad(0.25RS,$ $\quad(RS,$ |
| $S - x$ | |
| $min(\frac{1}{4x+\epsilon}, 1000)$ | |
| $\frac{1}{4x+\epsilon}$ | |
| $\min\left(\alpha R, \frac{1}{50x+\epsilon}\right)$ | $\alpha \in 0.5, 2, 4$ |
| $\max(\alpha RS, x)$ | $\alpha \in 0.4, 0.8$ |
| $\alpha Rx$ | $\alpha \in 2, 3$ |
| $\frac{x}{4\max(0.2,R)}$ | |
| $S^2 - x^2$ | |
| $(S - x)^2$ | |
| $x^{\alpha R}$ | $\alpha \in 1, 2$ |
| $S - x^{\max(\alpha R, 0.05)}$ | $\alpha \in 1, 2$ |
| $\begin{cases} 1, & \text{if } x > \max(R, 0.6)S \\ 0, & \text{otherwise} \end{cases}$ | |
| $\begin{cases} 1, & \text{if } x < \max(R, 0.4)S \\ 0, & \text{otherwise} \end{cases}$ | |
| $\begin{cases} 1, & \text{if } x < 0.25RS \text{ or } x > 0.75RS \\ 0, & \text{otherwise} \end{cases}$ | |
| $\begin{cases} 1, & \text{if } \max(0.25R, 0.1)S < x < \max(0.75R, 0.4)S \\ 0, & \text{otherwise} \end{cases}$ | |
| $x$ | |
| $x^2$ | |
| $\sqrt{x}$ | |
| $\sin(x) + 1$ | |
| $\cos(x) + 1$ | |
| $|\sin(x)|$ | |
| $|\cos(x)|$ | |
| $\left|\frac{\sin(x)}{x+\epsilon}\right|$ | |

$R$ is a uniform random number in the range $[0, 1]$: $R \in U([1, 0])$ and $S$ is the upper bound of the domain in the respective dimension

# Appendix 3: Tabulated data

In this section we present the tabulated data of the results depicted in Fig. 4 in Tables 3, 4, 5, 6, 7, 8 and 9. The synthetic results shown in Fig. 4 are the combined results from all synthetic test sets of the respective sample size and dimensionality. In addition to the density estimation methods reported in the main paper, we did additional tests which are reported in the tables. These are the smoothed cross-validation bandwidth estimator $R_{\text{scv}}$ [11, 20], the least squares cross-validation bandwidth estimator $R_{\text{lscv}}$ [5, 30] and a variational Bayesian Gaussian mixture model (BGMM) [1]. The methods are not included in the analysis in the main text, as it is easily visible from the main text that they are clearly worse in all aspects compared to their competing methods, which is GMM for BGMM and the

**Table 3** MSE, KL divergence, $p$ value and computing time for all evaluated methods in 1D, evaluated on the stock market dataset

| Estimator | $n = 500$ | | | | $n = 1000$ | | | |
|---|---|---|---|---|---|---|---|---|
| | MSE | KL Div. | $p$ value | Time (s) | MSE | KL Div. | $p$ value | Time (s) |
| KDE | 1.4e−1 | 2.4e−02 | 5.0e−01 | 1.6e+00 | 1.1e−01 | 2.2e−02 | 5.0e−01 | 4.5e+00 |
| $R_{\text{pi}}$ | 9.7e−02 | 1.9e−02 | 5.3e−01 | 3.5e−01 | 6.9e−02 | 1.5e−02 | 4.6e−01 | 3.6e−01 |
| $R_{\text{lscv}}$ | **6.8e−02** | **1.7e−02** | **7.0e−01** | 1.3e+01 | 4.5e−02 | **1.2e−02** | **7.1e−01** | 1.3e+01 |
| $R_{\text{scv}}$ | 9.7e−02 | 1.9e−02 | 5.2e−01 | 2.1e+00 | 6.8e−02 | 1.5e−02 | 4.8e−01 | 2.2e+00 |
| $R_{\text{nm}}$ | 1.9e−01 | 2.3e−02 | 6.2e−01 | 2.5e+02 | 8.6e−02 | 1.6e−02 | 7.0e−01 | 5.9e+02 |
| $R_{\text{ns}}$ | 1.8e−01 | 2.7e−02 | 3.2e−01 | **1.4e−01** | 1.4e−01 | 2.5e−02 | 2.3e−01 | **1.5e−01** |
| GMM | 2.1e+00 | – | – | 1.1e+03 | 3.3e+00 | – | – | 1.5e+03 |
| BGMM | 2.0e+01 | – | – | 8.3e+02 | 4.3e+01 | – | – | 1.9e+03 |
| MAF | 5.0e−01 | – | – | 2.9e+03 | 4.5e−01 | – | – | 3.0e+03 |
| FJE | 8.5e−02 | 1.8e−02 | 4.5e−01 | 2.6e+03 | 5.1e−02 | 1.3e−02 | 4.4e−01 | 3.5e+03 |
| DDE | 7.2e−02 | 1.8e−02 | 5.5e−01 | 1.1e+01 | 4.4e−02 | 1.3e−02 | 5.8e−01 | 2.1e+01 |
| DDE$_{\text{smooth}}$ | 7.1e−02 | 1.8e−02 | 5.4e−01 | 1.1e+01 | **4.1e−02** | **1.2e−02** | 5.7e−01 | 2.1e+01 |

| Estimator | $n = 5000$ | | | | $n = 10,000$ | | | |
|---|---|---|---|---|---|---|---|---|
| | MSE | KL Div. | $p$−value | Time (s) | MSE | KL Div. | $p$ value | Time (s) |
| KDE | 8.0e−02 | 1.5e−02 | 5.1e−01 | 5.8e+01 | 6.0e−02 | 1.3e−02 | 5.0e−01 | 2.1e+02 |
| $R_{\text{pi}}$ | 3.9e−02 | 8.3e−03 | 3.4e−01 | 4.5e−01 | 2.6e−02 | 6.1e−03 | 2.9e−01 | 5.8e−01 |
| $R_{\text{lscv}}$ | **2.0e−02** | **5.3e−03** | 7.2e−01 | 1.3e+00 | **1.3e−02** | **3.7e−03** | 7.2e−01 | 1.5e+00 |
| $R_{\text{scv}}$ | 3.8e−02 | 8.1e−03 | 3.0e−01 | 2.1e+00 | 2.6e−02 | 6.0e−03 | 3.0e−01 | 2.2e+00 |
| $R_{\text{nm}}$ | 3.4e−02 | 7.2e−03 | **7.4e−01** | 3.7e+03 | 1.6e−02 | 4.3e−03 | **7.3e−01** | 8.0e+03 |
| $R_{\text{ns}}$ | 9.6e−02 | 1.7e−02 | 5.2e−02 | **2.0e−01** | 7.6e−02 | 1.4e−02 | 2.0e−02 | **2.4e−01** |
| GMM | 2.0e+01 | – | – | 7.5e+03 | 4.2e+01 | – | – | 1.0e+04 |
| BGMM | 1.3e+04 | – | – | 1.2e+04 | 1.9e+02 | – | – | 2.3e+04 |
| MAF | 4.3e−01 | – | – | 3.1e+03 | 4.1e−01 | – | – | 4.1e+03 |
| FJE | 3.2e−02 | 7.7e−03 | 4.3e−01 | 6.7e+03 | 2.2e−02 | 5.7e−03 | 4.2e−01 | 8.5e+03 |
| DDE | 2.4e−02 | 7.8e−03 | 4.3e−01 | 2.2e+02 | 2.1e−02 | 6.9e−03 | 3.5e−01 | 5.4e+02 |
| DDE$_{\text{smooth}}$ | **2.0e−02** | 6.1e−03 | 4.3e−01 | 2.2e+02 | 1.8e−02 | 5.5e−03 | 3.4e−01 | 5.4e+02 |

**Table 4** MSE and computing time for KDE, $R_{nm}$, $R_{ns}$, and the smoothed DDE estimate in 1D, evaluated on synthetic test sets with different characteristics

| dataset | KDE | | $R_{nm}$ | | $R_{ns}$ | | DDE$_{\text{smooth}}$ | |
|---|---|---|---|---|---|---|---|---|
| | MSE | Time (s) | MSE | Time (s) | MSE | Time (s) | MSE | Time (s) |
| Gaussian 500 | 1.7e+0 | **3.6e−1** | 7.4e+1 | 1.1e+1 | 1.8e+1 | 4.1e−1 | **1.3e+0** | 2.1e+0 |
| Gaussian 1000 | 2.0e+0 | **1.0e+0** | 3.1e+1 | 2.1e+1 | 8.3e+0 | 1.6e+0 | **1.9e+0** | 3.1e+0 |
| Gaussian 5000 | 3.1e+0 | 1.7e+1 | **9.8e−1** | 1.8e+2 | 2.1e+1 | 4.2e+1 | 1.0e+5 | **1.3e+1** |
| Gaussian 10000 | **2.6e+0** | 6.3e+1 | 2.9e+2 | 4.0e+2 | 1.1e+2 | 1.8e+2 | 5.8e+3 | **5.4e+1** |
| Linear 500 | 5.7e−1 | **1.9e−1** | 3.2e−1 | 2.1e+1 | 1.6e−1 | 4.1e−1 | **8.3e−2** | 2.0e+0 |
| Linear 1000 | 6.7e−1 | **6.4e−1** | 1.7e−1 | 4.1e+1 | 2.3e−1 | 1.6e+0 | **7.6e−2** | 2.8e+0 |
| Linear 5000 | 5.8e−1 | **8.0e+0** | 3.4e−2 | 3.2e+2 | 6.7e−2 | 4.4e+1 | **3.0e−2** | 1.2e+1 |
| Linear 10000 | 6.0e−1 | 2.9e+1 | **2.3e−2** | 7.1e+2 | 9.5e−2 | 1.8e+2 | 3.4e−2 | **2.4e+1** |
| Monotone 500 | 4.9e−1 | **2.0e−1** | 1.6e−1 | 2.4e+1 | 1.1e−1 | 4.1e−1 | **4.6e−2** | 2.0e+0 |
| Monotone 1000 | 5.5e−1 | **6.0e−1** | 1.1e−1 | 4.9e+1 | 1.7e−1 | 1.6e+0 | **4.0e−2** | 2.9e+0 |
| Monotone 5000 | 5.7e−1 | **9.5e+0** | 5.5e−2 | 3.0e+2 | 9.0e−2 | 4.3e+1 | **2.7e−2** | 1.1e+1 |
| Monotone 10000 | 5.3e−1 | 2.9e+1 | **3.5e−2** | 7.9e+2 | 6.8e−2 | 1.8e+2 | 2.1e+2 | **2.7e+1** |
| Sinusoidal 500 | 5.2e−1 | **1.8e−1** | 5.3e−1 | 2.1e+1 | 2.6e−1 | 4.3e−1 | **5.0e−2** | 2.0e+0 |
| Sinusoidal 1000 | 5.8e−1 | **5.3e−1** | 3.0e−1 | 4.4e+1 | 1.2e−1 | 1.7e+0 | **3.4e−2** | 2.8e+0 |
| Sinusoidal 5000 | 5.4e−1 | **6.8e+0** | 1.1e−1 | 3.2e+2 | 8.1e−2 | 4.4e+1 | **2.0e−2** | 1.1e+1 |
| Sinusoidal 10000 | 5.6e−1 | 2.6e+1 | 3.4e−2 | 6.5e+2 | 5.2e−2 | 1.8e+2 | **2.3e−2** | **2.3e+1** |

**Table 5** MSE, KL divergence and computing time for all evaluated methods in 3D, evaluated on the deep lesion dataset

| Estimator | $n = 500$ | | | $n = 1000$ | | |
|---|---|---|---|---|---|---|
| | MSE | KL Div. | Time (s) | MSE | KL Div. | Time (s) |
| KDE | 4.6e+00 | 1.2e−01 | **7.5e+00** | 4.0e+00 | 1.2e−01 | 3.0e+01 |
| $R_{\text{pi}}$ | 1.0e+00 | 1.1e−01 | 1.8e+02 | **8.3e−01** | 9.3e−02 | 5.8e+02 |
| $R_{\text{lscv}}$ | **9.9e−01** | **1.0e−01** | 5.3e+03 | **8.3e−01** | 9.2e−02 | 2.1e+04 |
| $R_{\text{scv}}$ | 1.2e+00 | **1.0e−01** | 4.7e+03 | 9.9e−01 | **9.1e−02** | 1.7e+04 |
| $R_{\text{nm}}$ | 1.1e+00 | **1.0e−01** | 2.8e+03 | 8.5e−01 | 9.4e−02 | 6.7e+03 |
| $R_{\text{ns}}$ | 1.2e+00 | **1.0e−01** | 2.4e+01 | 1.1e+00 | 9.3e−02 | 5.7e+01 |
| GMM | 4.2e+00 | – | 3.0e+02 | 3.8e+00 | – | 4.2e+03 |
| BGMM | 4.3e+01 | – | 7.6e+02 | 8.3e+01 | – | 1.6e+03 |
| MAF | 1.1e+01 | – | 2.1e+03 | 3.3e+00 | – | 2.4e+03 |
| DDE | 1.4e+00 | 1.8e−01 | 9.9e+00 | 1.1e+00 | 1.5e−01 | **1.9e+01** |
| Estimator | $n = 5000$ | | | $n = 10,000$ | | |
| | MSE | KL Div. | Time (s) | MSE | KL Div. | Time (s) |
| KDE | 3.3e+00 | 1.1e−01 | 7.1e+02 | 3.0e+00 | 1.1e−01 | 3.0e+03 |
| $R_{\text{pi}}$ | 7.9e−01 | **7.3e−02** | 3.1e+03 | 7.3e−01 | 6.8e−02 | 2.2e+03 |
| $R_{\text{lscv}}$ | 1.7e+07 | 7.5e−02 | 1.3e+04 | – | – | – |
| $R_{\text{scv}}$ | 9.1e−01 | 7.7e−02 | 1.3e+04 | 8.2e−01 | 7.2e−02 | 1.1e+04 |
| $R_{\text{nm}}$ | **7.2e−01** | **7.3e−02** | 3.3e+04 | **6.8e−01** | **6.7e−02** | 1.0e+03 |
| $R_{\text{ns}}$ | 1.0e+00 | 8.0e−02 | **1.5e+01** | 9.3e−01 | 7.6e−02 | **1.5e+01** |
| GMM | 2.1e+01 | – | 1.6e+04 | 7.2e+01 | – | 2.3e+04 |
| BGMM | 3.2e+02 | – | 1.1e+04 | 4.8e+02 | – | 2.6e+04 |
| MAF | 1.9e+00 | – | 3.2e+03 | 1.9e+00 | – | 4.6e+03 |
| DDE | 8.4e−01 | 1.1e−01 | 1.1e+02 | 7.6e−01 | 9.9e−02 | 2.6e+02 |

**Table 6** MSE and computing time for KDE, $R_{nm}$, $R_{ns}$ and DDE in 3D, evaluated on synthetic test sets with different characteristics

| Data-set | KDE | | $R_{nm}$ | | $R_{ns}$ | | DDE | |
|---|---|---|---|---|---|---|---|---|
| | MSE | Time (s) | MSE | Time (s) | MSE | Time (s) | MSE | Time (s) |
| Gaussian 500 | 1.1e+1 | **7.1e−1** | **4.5e+0** | 4.1e+2 | 9.6e+0 | 4.1e+0 | 9.1e+0 | 2.0e+0 |
| Gaussian 1000 | 2.4e+1 | 2.8e+0 | **1.4e+1** | 8.4e+2 | 2.2e+1 | 9.5e+0 | 1.9e+1 | **2.6e+0** |
| Gaussian 5000 | 1.6e+1 | 5.0e+1 | **4.9e+0** | 3.0e+3 | 1.7e+1 | 9.2e+1 | 1.1e+1 | **8.9e+0** |
| Gaussian 10000 | 6.5e+0 | 1.7e+2 | **1.8e+0** | 5.5e+3 | 7.8e+0 | 3.0e+2 | 4.0e+0 | **2.2e+1** |
| Linear 500 | 3.6e−1 | **7.0e−1** | 4.3e−1 | 3.4e+2 | 2.9e−1 | 4.1e+0 | **2.5e−1** | 1.8e+0 |
| Linear 1000 | 2.7e−1 | 2.7e+0 | 2.9e−1 | 8.0e+2 | 2.7e−1 | 9.4e+0 | **2.0e−1** | 2.5e+0 |
| Linear 5000 | 2.7e−1 | 4.8e+1 | – | – | 3.5e−1 | 9.1e+1 | **2.3e−1** | 8.5e+0 |
| Linear 10000 | 2.1e−1 | 1.6e+2 | **1.6e−1** | 1.3e+2 | 3.0e−1 | 2.9e+2 | 2.0e−1 | **1.7e+1** |
| Linear 50000 | **1.3e−1** | 3.7e+3 | – | – | – | – | 1.5e−1 | **8.4e+1** |
| Monotone 500 | 6.2e−1 | **7.0e−1** | **4.9e−1** | 3.6e+2 | 5.3e−1 | 4.1e+0 | **4.9e−1** | 1.9e+0 |
| Monotone 1000 | 5.8e−1 | 2.8e+0 | 5.4e−1 | 8.4e+2 | 5.0e−1 | 9.2e+0 | **4.4e−1** | **2.5e+0** |
| Monotone 5000 | 2.3e−1 | 4.8e+1 | – | – | 2.7e−1 | 9.1e+1 | **1.9e−1** | **8.3e+0** |
| Monotone 10000 | 1.9e−1 | 1.7e+2 | **1.5e−1** | 7.5e+3 | 2.5e−1 | 2.9e+2 | 1.8e−1 | **1.6e+1** |
| Monotone 50000 | – | – | – | – | – | – | **2.2e−1** | **8.6e+1** |
| Sinusoidal 500 | 2.4e−1 | **6.9e−1** | 2.1e−1 | 3.5e+2 | 2.2e−1 | 4.1e+0 | **1.5e−1** | 1.9e+0 |
| Sinusoidal 1000 | 2.1e−1 | 2.7e+0 | 1.9e−1 | 6.9e+2 | 2.4e−1 | 9.4e+0 | **1.4e−1** | **2.5e+0** |
| Sinusoidal 5000 | 1.0e−1 | 4.6e+1 | – | – | 1.7e−1 | 9.1e+1 | **8.5e−2** | **8.6e+0** |
| Sinusoidal 10000 | 8.3e−2 | 1.6e+2 | 8.0e−2 | 1.0e+2 | 1.5e−1 | 2.9e+2 | **7.8e−2** | **1.6e+1** |
| Sinusoidal 50000 | – | – | – | – | – | – | **6.9e−2** | **8.5e+1** |

Table 7 MSE, KL divergence and computing time for all evaluated methods in 5D

| Dataset | KDE | | | $R_{um}$ | | | $R_{ss}$ | | | GMM | | | BGMM | | | MAF | | | DDE | | |
|---|---|---|---|---|---|---|---|---|---|---|---|---|---|---|---|---|---|---|---|---|---|
| | MSE | KL div. | Time (s) | MSE | KL div. | Time (s) | MSE | KL div. | Time (s) | MSE | KL div. | Time (s) | MSE | KL div. | Time (s) | MSE | KL div. | Time (s) | MSE | KL div. | Time (s) |
| Gaussian 500 | 1.2e+1 | 3.6e−1 | **7.3e−1** | 2.7e+2 | **2.8e−1** | 1.4e+3 | **7.9e+1** | 3.1e−1 | 4.9e+0 | 1.0e−1 | – | 2.9e+2 | 2.8e+2 | – | 2.9e+2 | 9.8e+1 | – | 7.5e+1 | 8.0e+0 | 3.1e−1 | 2.0e+0 |
| Gaussian 1000 | 1.5e+1 | 4.0e−1 | **2.9e+0** | 1.4e+2 | **3.2e−1** | 2.6e+3 | **1.1e+1** | 3.3e−1 | 1.2e+1 | 1.2e−1 | – | 5.8e+2 | 3.3e+2 | – | 5.8e+2 | 1.3e+1 | – | 1.5e+2 | **1.1e+1** | 3.4e−1 | 3.0e+0 |
| Gaussian 5000 | 1.1e+1 | 3.4e−1 | 7.5e+1 | 2.3e+1 | **2.6e−1** | 9.5e+1 | 8.5e+1 | 2.7e−1 | 1.2e+2 | 3.6e+1 | – | 3.2e+2 | 5.9e+2 | – | 3.2e+3 | 1.1e+1 | – | 9.8e+2 | **7.6e+0** | 2.7e−1 | **1.2e+1** |
| Gaussian 10000 | 1.1e+1 | 3.0e−1 | 3.2e+2 | **6.1e+0** | **2.0e−1** | 1.5e+2 | 8.2e+0 | 2.3e−1 | 4.1e+2 | 1.2e+2 | – | 4.6e+3 | 6.3e+2 | – | 4.6e+3 | 1.1e+1 | – | 2.5e+3 | 7.2e+0 | 2.3e−1 | **3.5e+1** |
| Linear 500 | 1.6e+0 | 1.2e−1 | **7.2e−1** | 2.2e+0 | 8.8e−2 | 6.7e+2 | 4.5e−1 | 9.2e−2 | 4.8e+0 | 2.1e+0 | – | 2.8e+2 | 2.9e+2 | – | 2.8e+2 | 1.3e+0 | – | 7.4e+1 | 4.4e−1 | **8.0e−1** | 1.9e+0 |
| Linear 1000 | 1.5e+0 | 1.2e−1 | **2.9e+0** | 2.7e+0 | 9.2e−2 | 1.7e+3 | 4.6e−1 | 8.8e−2 | 1.1e+1 | 1.9e−1 | – | 5.5e+2 | 3.7e+2 | – | 5.5e+2 | 1.3e+0 | – | 1.7e+2 | **4.1e−1** | 7.5e−2 | **2.9e+0** |
| Linear 5000 | 1.1e+0 | 1.1e−1 | 7.4e+1 | 9.0e−1 | 7.6e−2 | 8.5e+1 | 4.7e−1 | 7.4e−2 | 1.2e+2 | 2.5e+1 | – | 3.4e+3 | 6.3e+2 | – | 3.4e+3 | 1.3e+0 | – | 1.2e+3 | **3.4e−1** | **6.3e−2** | **1.3e+1** |
| Linear 10000 | 1.0e+0 | 1.0e−1 | 3.1e+2 | 5.5e−1 | 7.1e−2 | 1.4e+2 | 4.9e−1 | 7.2e−2 | 4.0e+2 | 1.2e−2 | – | 5.0e+3 | 6.7e+2 | – | 5.0e+3 | 1.4e+0 | – | 2.9e+3 | **3.4e−1** | **6.2e−2** | **2.5e+1** |
| Monotone 500 | 1.6e+0 | 1.1e−1 | **7.1e−1** | 1.1e+0 | 8.4e−2 | 5.8e+2 | 6.7e−1 | 8.8e−2 | 4.8e+0 | 2.0e+0 | – | 2.8e+2 | 2.5e+2 | – | 2.8e+2 | 1.4e+0 | – | 7.9e+1 | **6.4e−1** | 7.5e−2 | 1.9e+0 |
| Monotone 1000 | 1.2e+0 | 1.0e−1 | **2.9e+0** | 1.9e+0 | 7.3e−2 | 1.5e+3 | 3.7e−1 | 7.6e−2 | 1.1e+1 | 1.4e+0 | – | 5.4e+2 | 3.6e+2 | – | 5.4e+2 | 1.1e+0 | – | 1.5e+2 | **3.2e−1** | **6.0e−2** | 2.8e+0 |
| Monotone 5000 | 9.2e−1 | 9.8e−2 | 7.4e+1 | 9.0e−1 | 6.8e−2 | 8.0e+1 | 4.6e−1 | 7.1e−2 | 1.2e+2 | 2.1e+1 | – | 3.4e+3 | 6.1e+2 | – | 3.4e+3 | 1.2e+0 | – | 1.3e+3 | **3.7e−1** | **5.6e−2** | **1.2e+1** |
| Monotone 10000 | 8.0e−1 | 9.0e−2 | 3.1e+2 | 4.8e−1 | 6.0e−2 | 1.4e+2 | 3.9e−1 | 6.4e−2 | 4.1e+2 | 1.0e+0 | – | 5.0e+3 | 6.7e+2 | – | 5.0e+3 | 1.2e+0 | – | 2.7e+3 | **2.9e−1** | **5.0e−2** | **2.5e+1** |
| Sinusoidal 500 | 7.6e−1 | 1.1e−1 | **7.1e−1** | 7.7e−1 | **5.6e−2** | 7.5e+2 | 1.7e−1 | 7.0e−2 | 4.9e+2 | 1.1e+0 | – | 2.7e+2 | 2.9e+2 | – | 2.7e+2 | 6.8e−1 | – | 7.2e+1 | **1.5e−1** | 5.8e−2 | 2.0e+0 |
| Sinusoidal 1000 | 6.7e−1 | 1.0e−1 | **2.9e+0** | 1.3e+0 | 4.9e−2 | 1.4e+3 | 1.7e−1 | 6.5e−2 | 1.1e+1 | 8.7e−1 | – | 5.3e+2 | 4.0e+2 | – | 5.3e+2 | 6.7e−1 | – | 1.5e+2 | **1.3e−1** | **4.5e−1** | **2.9e+0** |
| Sinusoidal 5000 | 5.2e−1 | 1.0e−1 | 7.4e+1 | 4.4e−1 | 5.1e−2 | 4.2e+3 | 2.1e−1 | 6.7e−2 | 1.2e+2 | 2.8e+1 | – | 3.1e+3 | 6.5e+2 | – | 3.1e+2 | 7.3e−1 | – | 1.3e+3 | **1.2e−1** | 3.9e−1 | **1.1e+1** |
| Sinusoidal 10000 | 4.0e−1 | 8.1e−2 | 3.1e+2 | 2.4e−1 | 4.4e−2 | 1.3e+2 | 1.7e−1 | 5.5e−2 | 4.1e+2 | 9.5e+1 | – | 4.6e+3 | 6.8e+2 | – | 4.6e+3 | 6.9e−1 | – | 2.9e+3 | **1.1e−1** | 3.1e−1 | **2.5e+1** |

**Table 8** MSE, KL divergence and computing time for all evaluated methods in 10D

| Dataset | KDE | | | $R_{nn}$ | | | $R_{ss}$ | | | GMM | | | BGMM | | | MAF | | | DDE | | |
|---|---|---|---|---|---|---|---|---|---|---|---|---|---|---|---|---|---|---|---|---|---|
| | MSE | KL div. | Time (s) | MSE | KL div. | Time (s) | MSE | KL div. | Time (s) | MSE | KL div. | Time (s) | MSE | KL div. | Time (s) | MSE | KL div. | Time (s) | MSE | KL div. | Time (s) |
| Gaussian 500 | 6.2e+0 | 3.0e−1 | **8.1e−1** | 5.8e+2 | 2.8e−2 | 7.0e+3 | 9.6e+1 | 2.8e−1 | 2.9e+1 | 6.2e+0 | – | 2.1e+2 | 6.9e+2 | – | 2.1e+2 | 6.5e+0 | – | 1.2e+1 | 3.0e+0 | 2.7e−1 | 2.2e+0 |
| Gaussian 1000 | 7.5e+0 | 3.3e−1 | **3.3e+0** | 3.2e+3 | 3.0e−1 | 3.4e+4 | 5.2e+1 | 3.0e−1 | 5.9e+1 | 7.5e+0 | – | 1.1e+3 | 6.9e+2 | – | 1.1e+3 | 1.6e+1 | – | 6.0e+1 | 3.9e+0 | 2.9e−1 | 3.5e+0 |
| Gaussian 5000 | 6.3e+0 | 3.2e−1 | 8.5e+1 | – | – | – | 1.7e+1 | 2.7e−1 | 6.1e+2 | 4.9e+0 | – | 9.8e+3 | 6.4e+2 | – | 9.8e+3 | 5.1e+0 | – | 2.9e+3 | 3.1e+0 | 2.6e−1 | **3.0e+1** |
| Gaussian 10000 | 6.3e+0 | 3.3e−1 | 3.5e+2 | – | – | – | 1.1e+1 | 2.7e−1 | 1.9e+3 | 1.5e+1 | – | 1.5e+4 | 6.6e+2 | – | 1.5e+4 | 5.3e+0 | – | 5.8e+3 | 3.1e+0 | 2.6e−1 | **9.9e+1** |
| Linear 500 | 2.1e+0 | 1.2e−1 | **8.0e−1** | 1.3e+2 | 9.3e−2 | 2.3e+3 | 8.8e+1 | 9.3e−2 | 1.7e+1 | 2.1e+0 | – | 1.2e+2 | – | – | 1.2e+2 | 4.6e+0 | – | 6.9e+1 | 4.0e+0 | 8.6e−2 | 2.2e+0 |
| Linear 1000 | 2.1e+0 | 1.3e−1 | **3.2e+0** | 1.4e+2 | 9.9e−2 | 1.1e+4 | 5.4e+1 | 9.9e−2 | 3.8e+1 | 2.0e+0 | – | 1.1e+3 | 7.2e+2 | – | 1.1e+3 | 2.5e+0 | – | 6.6e+1 | 3.9e+0 | 8.9e−2 | 3.3e+0 |
| Linear 5000 | 1.8e+0 | 1.4e−1 | 8.4e+1 | – | – | – | 1.7e+1 | 7.9e−2 | 4.6e+2 | 1.3e+0 | – | 9.9e+3 | 6.5e+2 | – | 9.9e+3 | 1.4e+0 | – | 2.8e+3 | 3.0e+0 | 7.3e−2 | **3.0e+1** |
| Linear 10000 | 2.0e+0 | 1.6e−1 | 3.4e+2 | – | – | – | 1.1e+1 | 8.7e−2 | 1.5e+3 | 1.5e+1 | – | 1.6e+4 | 7.0e+2 | – | 1.6e+4 | 1.6e+0 | – | 6.5e+3 | 3.5e−1 | 7.8e−1 | **9.3e+1** |
| Monotone 500 | 3.5e+0 | 1.3e−1 | **8.1e−1** | 1.5e+2 | 1.1e−1 | 2.8e+3 | 1.1e+2 | 1.1e−1 | 1.6e+1 | 3.5e+0 | – | 1.2e+2 | – | – | 1.2e+2 | 5.9e+0 | – | – | 1.7e+0 | 1.0e−1 | 6.6e+0 |
| Monotone 1000 | 2.9e+0 | 1.3e−1 | **3.2e+0** | 1.4e+2 | 1.0e−1 | 2.2e+2 | 6.5e+1 | 1.0e−1 | 3.9e+1 | 3.0e+0 | – | 1.1e+3 | 7.2e+2 | – | 1.1e+3 | 2.9e+0 | – | 6.3e+1 | 1.3e+0 | 9.4e−2 | 3.4e+0 |
| Monotone 5000 | 8.2e+0 | 2.0e−1 | 8.4e+1 | – | – | – | 2.7e+1 | 1.5e−1 | 4.6e+2 | 7.1e+0 | – | 9.9e+3 | 6.3e+2 | – | 9.9e+3 | 7.7e+0 | – | 2.9e+3 | 6.3e+0 | 1.4e−1 | **3.0e+1** |
| Monotone 10000 | 6.4e+0 | 2.0e−1 | 3.4e+2 | – | – | – | 1.6e+1 | 1.4e−1 | 1.5e+3 | 1.7e+1 | – | 1.6e+4 | 6.8e+2 | – | 1.6e+4 | 5.8e+0 | – | 6.2e+3 | 4.3e+0 | 1.3e−1 | **9.3e+1** |
| Simsoidal 500 | 1.1e+0 | 7.5e−2 | **8.1e−1** | 8.5e+1 | **2.3e−2** | 2.0e+3 | 8.0e+1 | **2.3e−2** | 1.8e+1 | 1.2e+0 | – | 1.2e+2 | – | – | 1.2e+2 | 4.2e+0 | – | 6.4e+1 | 1.5e−1 | 3.6e−2 | 2.2e+0 |
| Simsoidal 1000 | 1.1e+0 | 8.4e−2 | **3.2e+0** | 8.5e+1 | **2.2e−2** | 1.3e+4 | 4.6e+1 | 2.3e−2 | 3.8e+1 | 1.1e+0 | – | 1.1e+3 | 7.3e+2 | – | 1.1e+3 | 1.8e+0 | – | 6.4e+1 | 1.2e−1 | 3.3e−2 | 3.4e+0 |
| Simsoidal 5000 | 1.1e+0 | 1.1e−1 | 8.3e+1 | – | – | – | 1.6e+1 | **2.1e−2** | 4.5e+2 | 1.2e+0 | – | 9.6e+3 | 6.5e+2 | – | 9.6e+3 | 9.3e−1 | – | 3.0e+3 | 9.9e−2 | 2.7e−2 | **3.0e+1** |
| Simsoidal 10000 | 1.1e+0 | 1.3e−1 | 3.4e+2 | – | – | – | 9.5e+0 | **2.4e−2** | 1.5e+3 | 1.3e+1 | – | 1.5e+4 | 7.0e+2 | – | 1.5e+4 | 9.4e−1 | – | 6.5e+3 | 1.0e−1 | 2.9e−2 | **9.4e+1** |

**Table 9** MSE, KL divergence and computing time for all evaluated methods in 30D

| Dataset | KDE | | | $R_{nx}$ | | | GMM | | | BGMM | | | MAF | | | DDE | | |
|---|---|---|---|---|---|---|---|---|---|---|---|---|---|---|---|---|---|---|
| | MSE | KL div. | Time (s) | MSE | KL div. | Time (s) | MSE | KL div. | Time (s) | MSE | KL div. | Time (s) | MSE | KL div. | Time (s) | MSE | KL div. | Time (s) |
| Gaussian 500 | 2.6e+0 | 1.7e−1 | 1.2e+0 | 6.8e+9 | **1.4e−1** | **1.2e+0** | 2.6e+0 | – | 3.9e+2 | 7.1e+2 | – | 1.1e+1 | 2.2e+13 | – | 1.2e+2 | **7.6e−1** | **1.4e−1** | 2.4e−1 |
| Gaussian 1000 | 2.6e+0 | 1.7e−1 | 4.8e+0 | 3.7e+9 | **1.4e−1** | **4.8e+0** | 2.6e+0 | – | 1.0e+3 | 7.3e+2 | – | 2.0e+1 | 7.4e+4 | – | 1.3e+2 | **7.8e−1** | **1.4e−1** | 4.9e+0 |
| Gaussian 5000 | 2.3e+0 | 1.7e−1 | 1.2e+2 | 1.8e+9 | **1.3e−1** | 1.2e+2 | 2.3e+0 | – | 1.7e+4 | 7.2e+2 | – | 1.3e+3 | 1.4e+1 | – | 2.0e+2 | **5.9e−1** | **1.3e−1** | **6.3e+1** |
| Gaussian 10000 | 2.7e+0 | 1.9e−1 | 4.9e+2 | 1.5e+9 | **1.5e−1** | 4.9e+2 | 2.7e+0 | – | 3.6e+4 | 6.5e+2 | – | 2.5e+3 | 3.0e+0 | – | 3.4e+2 | **8.3e−1** | **1.5e−1** | 2.3e+2 |
| Linear 500 | 1.5e+0 | 9.2e−2 | 1.2e+0 | 5.9e+9 | **6.1e−2** | **1.2e+0** | 1.5e+0 | – | 1.6e+2 | – | – | – | 4.4e+8 | – | 7.3e+1 | **2.2e−1** | **6.1e−2** | 2.4e+0 |
| Linear 1000 | 1.5e+0 | 9.0e−2 | 4.7e+0 | 3.1e+9 | **5.3e−2** | **4.7e+0** | 1.5e+0 | – | 3.4e+2 | 7.6e+2 | – | 1.4e+3 | 2.8e+5 | – | 8.3e+1 | **2.1e−1** | **5.3e−2** | 4.8e+0 |
| Linear 5000 | 1.6e+0 | 1.1e−1 | 1.2e+2 | 1.5e+9 | 6.8e−2 | 1.2e+2 | 1.6e+0 | – | 1.7e+4 | – | – | – | 8.1e+2 | – | 2.0e+2 | **2.6e−1** | **6.7e−2** | **6.0e+1** |
| Linear 10000 | 1.6e+0 | 1.1e−1 | 4.8e+2 | 1.2e+9 | **6.4e−2** | 4.8e+2 | 1.6e+0 | – | 2.6e+4 | – | – | – | 6.1e+0 | – | 4.5e+2 | **2.5e−1** | **6.4e−2** | 2.3e+2 |
| Monotone 500 | 2.3e+0 | 1.1e−1 | 1.1e+0 | 7.5e+9 | 8.3e−2 | 1.1e+0 | 2.3e+0 | – | 1.6e+2 | – | – | – | 6.6e+7 | – | 7.3e+1 | **8.1e−1** | **8.2e−2** | 2.4e+0 |
| Monotone 1000 | 5.5e+0 | 1.5e−1 | 4.7e+0 | 3.6e+9 | **1.2e−1** | 4.7e+0 | 5.5e+0 | – | 3.5e+2 | 7.5e+2 | – | 1.4e+3 | 3.3e+4 | – | 8.2e+1 | **3.9e+0** | **1.2e−1** | 4.9e+0 |
| Monotone 5000 | 6.4e+0 | 2.0e−1 | 1.2e+2 | 1.7e+9 | **1.6e−1** | 1.2e+2 | 6.4e+0 | – | 1.7e+4 | – | – | – | 2.2e+1 | – | 2.1e+1 | **4.7e+0** | **1.6e−1** | **6.3e+1** |
| Monotone 10000 | 6.5e+0 | 1.9e−1 | 4.8e+2 | 1.4e+9 | **1.4e−1** | 4.8e+2 | 6.5e+0 | – | 2.5e+4 | – | – | – | 7.5e+0 | – | 4.5e+2 | **4.9e+0** | **1.4e−1** | 2.3e+2 |
| Sinusoidal 500 | 1.0e+0 | 4.6e−2 | 1.2e+0 | 5.8e+9 | **7.6e−3** | **1.2e+0** | 1.0e+0 | – | 1.8e+2 | – | – | – | 2.4e+6 | – | 7.3e+1 | **7.1e−2** | 9.9e−3 | 2.4e+0 |
| Sinusoidal 1000 | 1.0e+0 | 4.9e−2 | 4.8e+0 | 3.0e+9 | **7.6e−3** | **4.8e+0** | 1.0e+0 | – | 3.7e+2 | 6.8e+2 | – | 1.4e+3 | 1.5e+4 | – | 8.3e+1 | **6.2e−2** | 9.6e−3 | **4.8e+0** |
| Sinusoidal 5000 | 1.0e+0 | 6.1e−2 | 1.2e+2 | 1.4e+9 | **7.4e−3** | 1.2e+2 | 1.0e+0 | – | 1.7e+4 | – | – | – | 9.8e+0 | – | 2.1e+2 | **5.4e−2** | 9.3e−3 | **6.1e+1** |
| Sinusoidal 10000 | 1.0e+0 | 6.3e−2 | 4.8e+2 | 1.2e+9 | **7.5e−3** | 4.8e+2 | 1.0e+0 | – | 2.6e+4 | – | – | – | 1.6e+0 | – | 4.6e+2 | **5.2e−2** | 9.0e−3 | 2.3e+2 |

other KDE estimators for $R_{\mathrm{scv}}$. $R_{\mathrm{lscv}}$ however is not included in the analysis for a different reason. First of all, it is very slow and thus only applicable for small domain dimensionalities, and secondly, even though it gives the best predictions sometimes, it also fails incredibly in other cases without an apparent reason. This causes an error many magnitudes larger than of all other methods. Such failure without any apparent reason is a clear exclusion criterion. For the reason of long computing times, not all methods are tested for all datasets. All tables contain the different estimators in columns, with MSE, KL divergence and computing time for each method and dataset reported in the main paper and with only the MSE and computing time for the other methods and/or datasets. In 1D we additionally report the p value, which is only applicable there. Every entry in the tables is the mean (or sum for the computing time) of the respective metric over the respective dataset for the respective method. The values highlighted in bold font are the best respective values for that dataset. The dataset names are combinations of the characteristic details of the included distributions and the number of points per distribution. The former is either the name of the real-world dataset (Stock data, Imagenet or DeepLesion) or the sole characteristic of the base functions from which the synthetic functions were generated (Gaussian, linear, monotone or sinusoidal). The real-world datasets contained 500 samples each, and the synthetic datasets contained 50 samples each. The synthetic functions with same characteristic but different sample sizes were all randomly generated anew and thus contain different ground truth densities.

## Appendix 4: Additional plots of the 1D analysis

In this section we present the plots of the PDFs and estimates of Sect. 5 individually in Fig. 16, 17, 18, 19, 20, 21, 22, 23, 24, 25 such that similar estimates can be better compared.

**Fig. 16** Estimates for the gamma distribution with $n = 500$

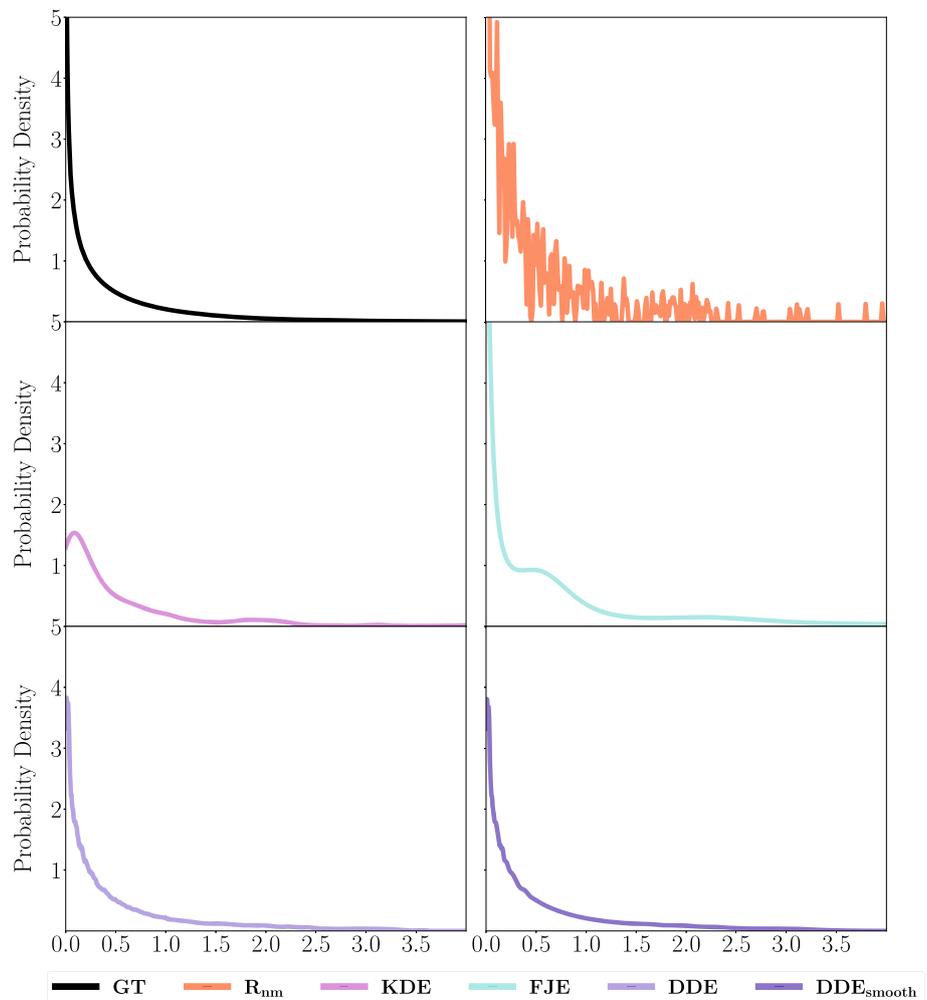

**Fig. 17** Estimates for the sum of two Gaussians distribution with $n = 500$

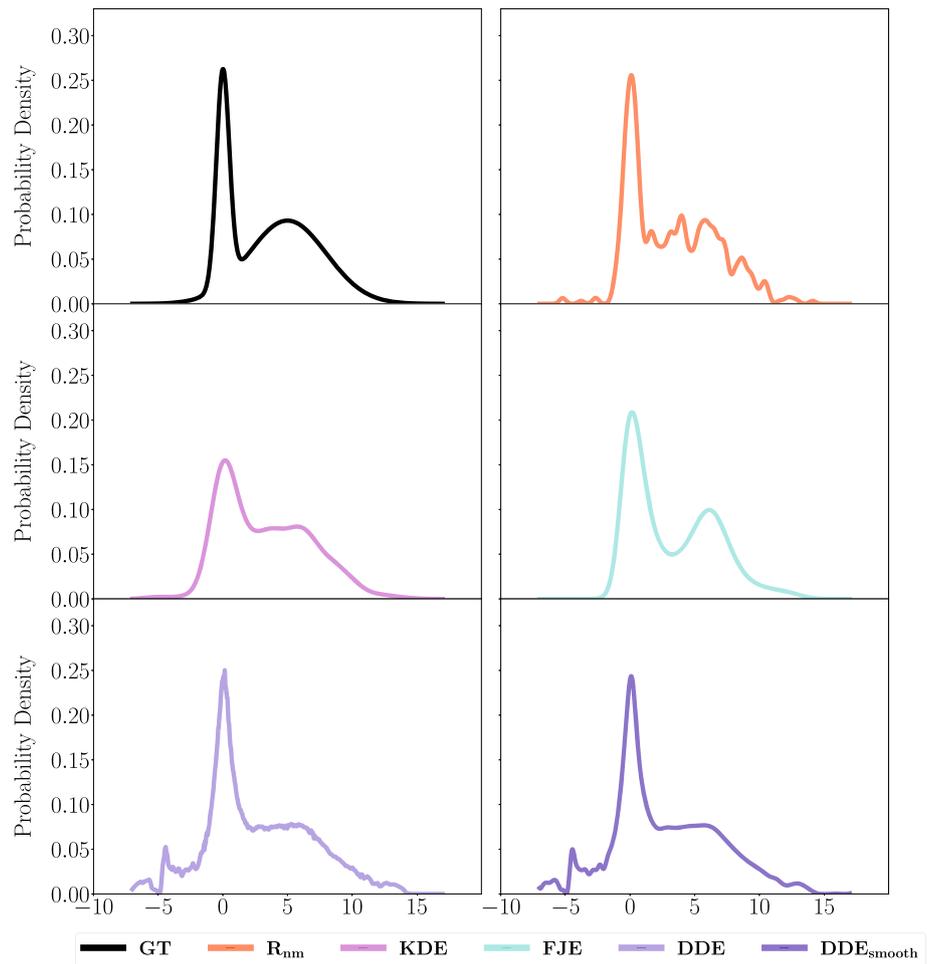

**Fig. 18** Estimates for the five fingers distribution with $n = 500$

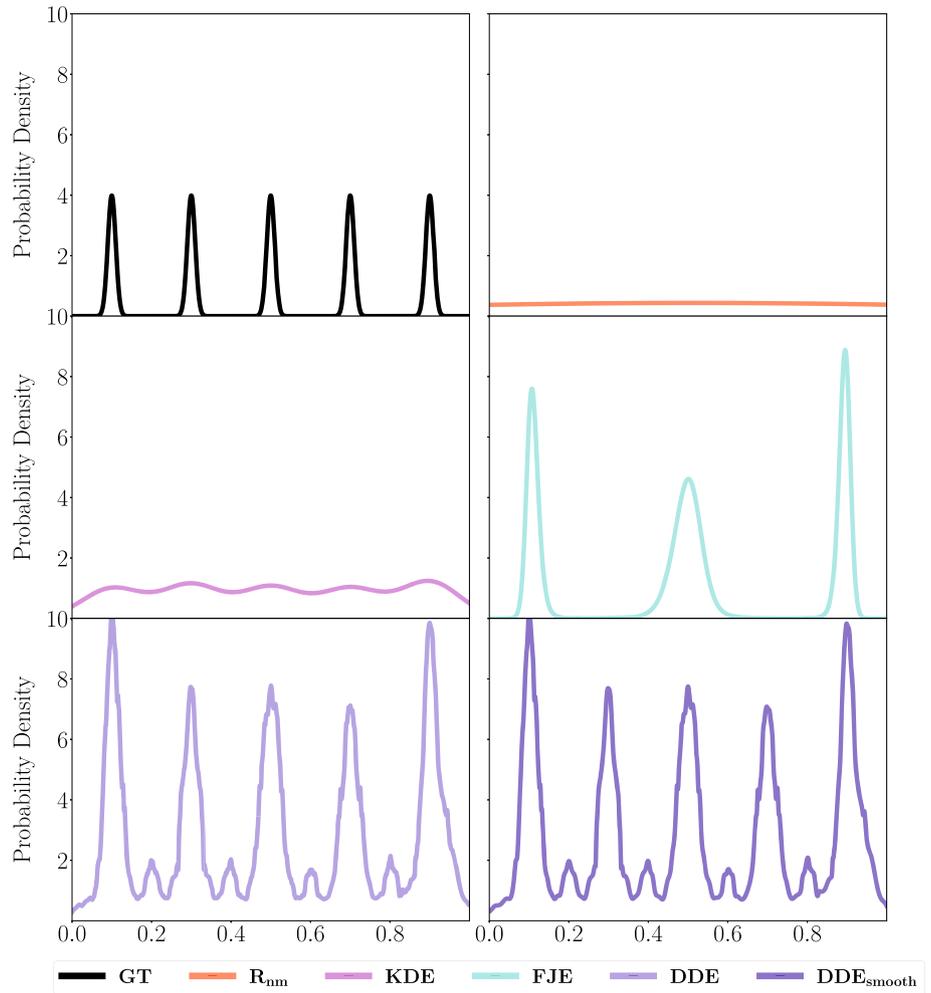

**Fig. 19** Estimates for the
Cauchy distribution with
$n = 500$

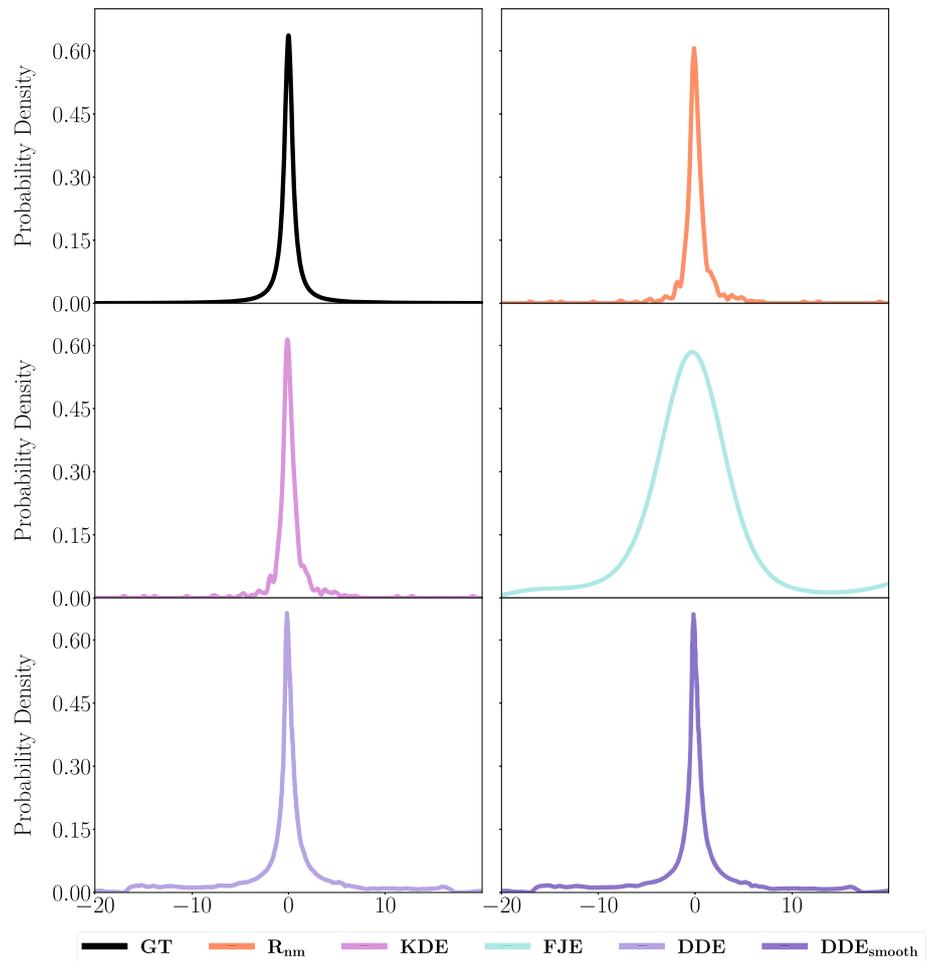

**Fig. 20** Estimates for the discontinuous distribution with $n = 500$

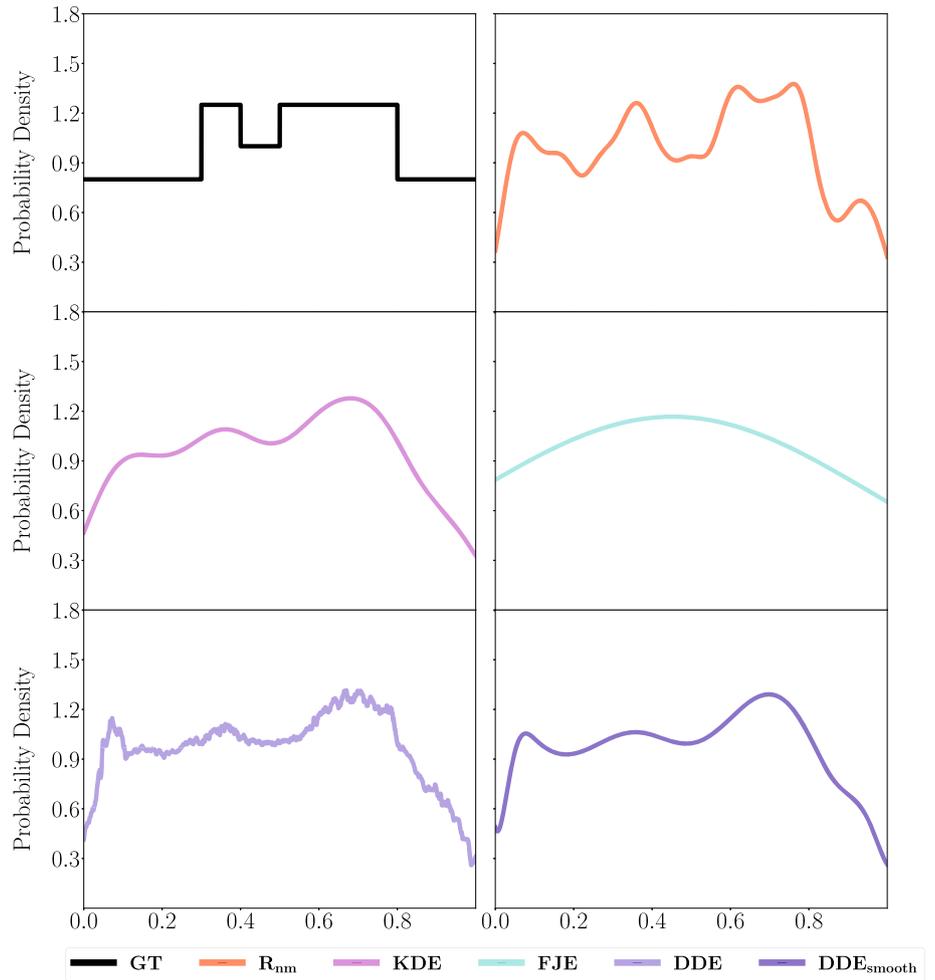

**Fig. 21** Estimates for the gamma distribution with $n = 5000$

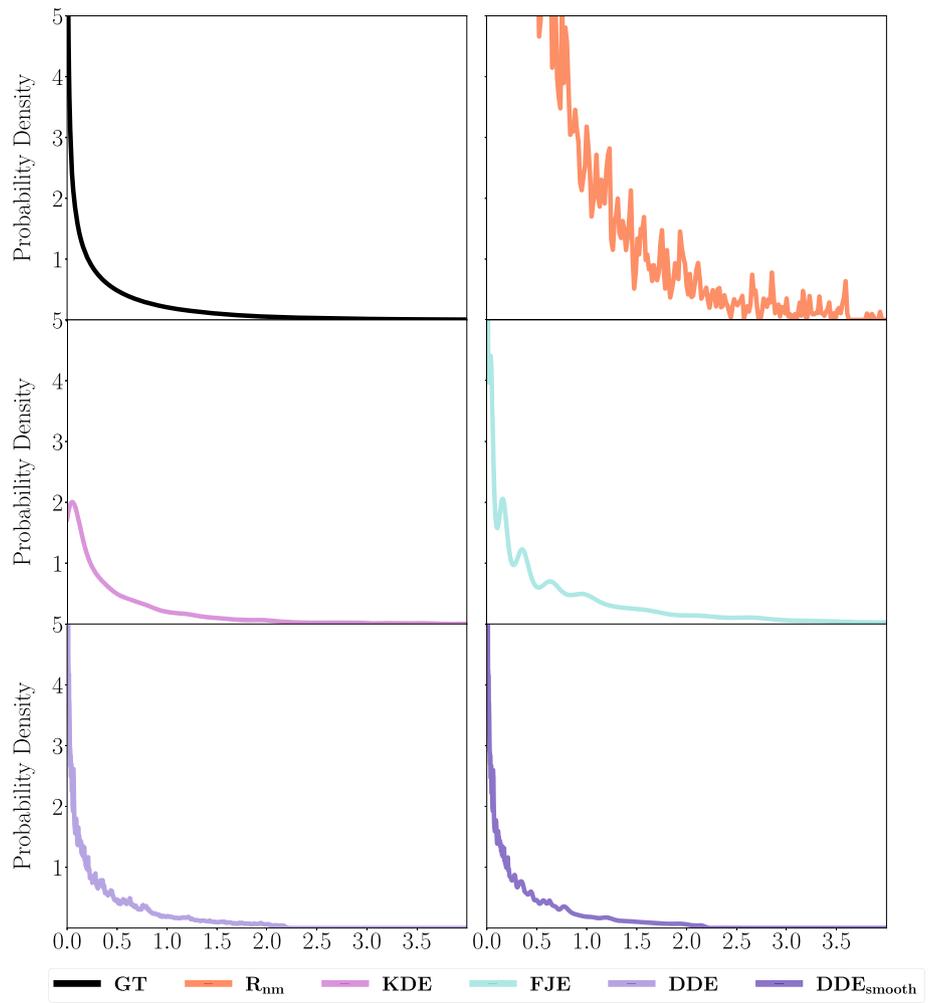

**Fig. 22** Estimates for the sum of two Gaussians distribution with $n = 5000$

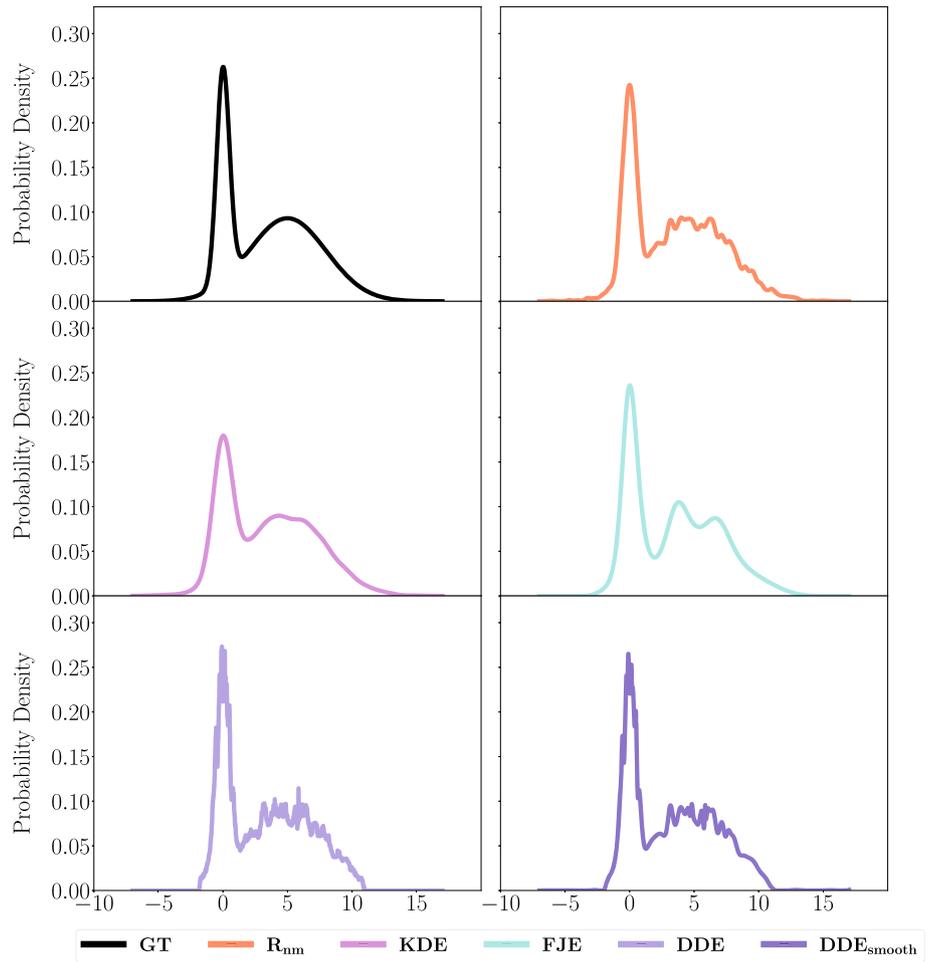

**Fig. 23** Estimates for the five fingers distribution with $n = 5000$

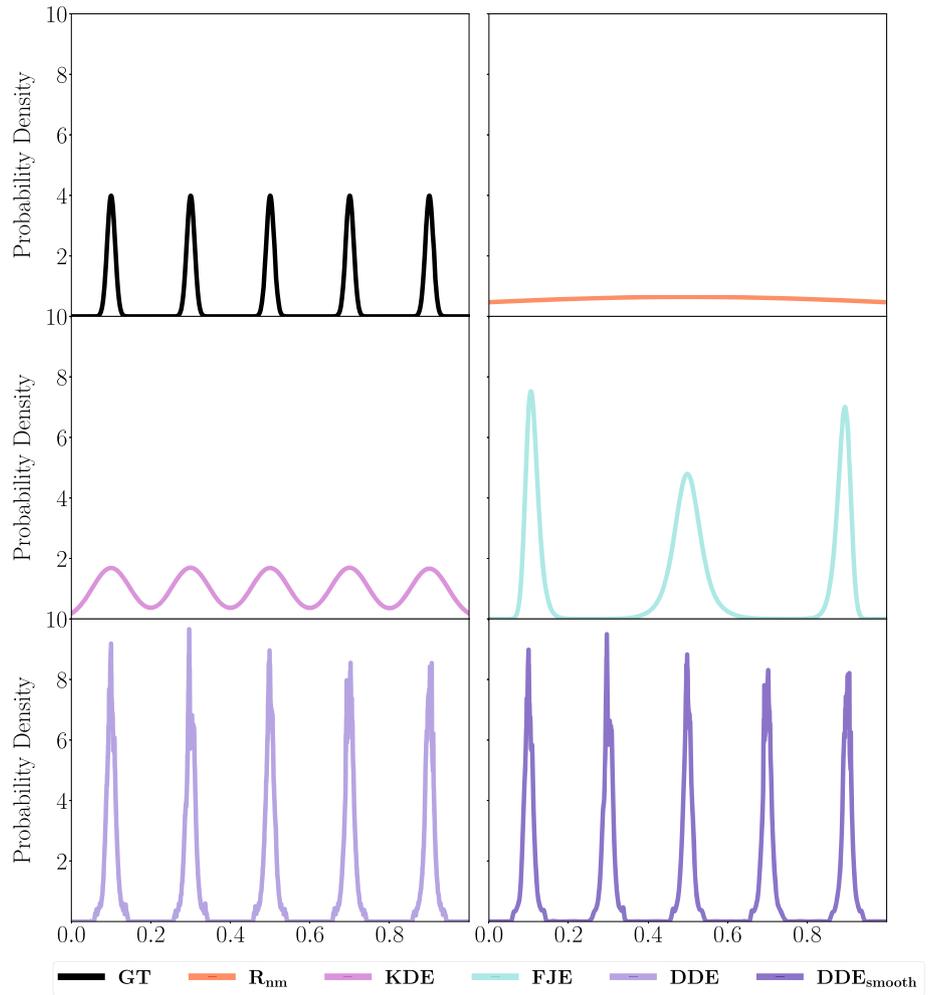

**Fig. 24** Estimates for the
Cauchy distribution with
$n = 5000$

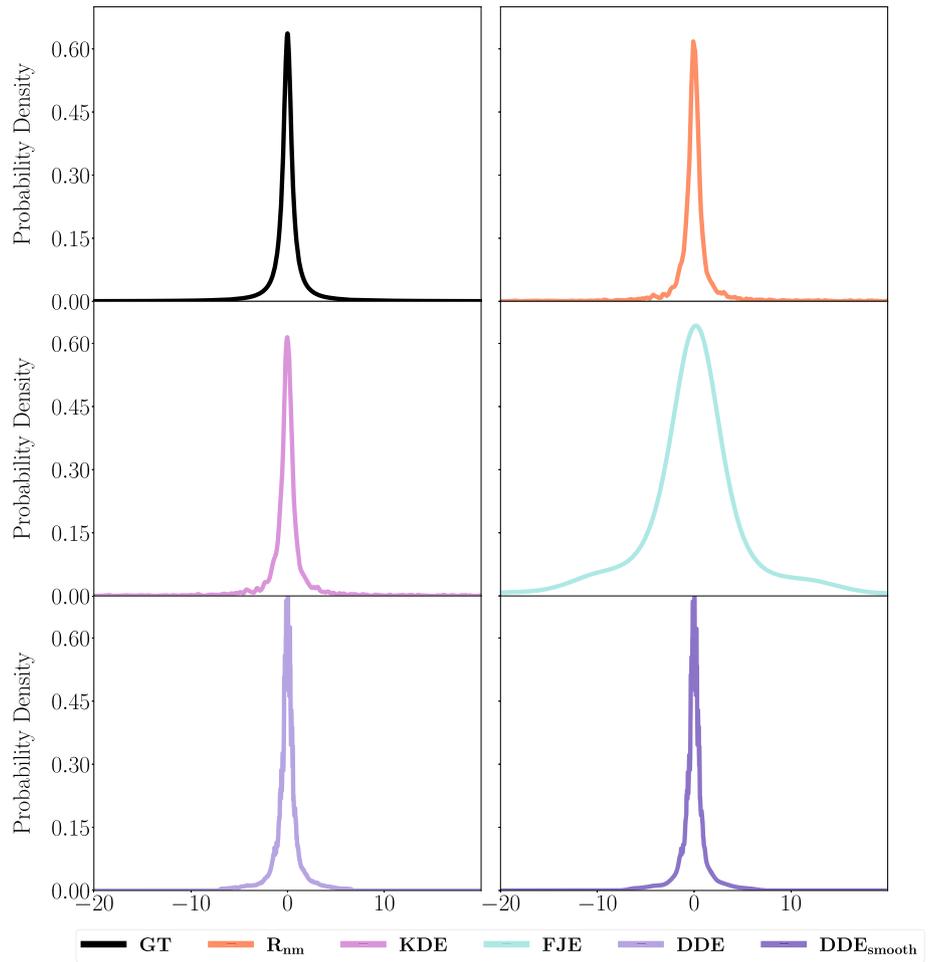

**Fig. 25** Estimates for the discontinuous distribution with $n = 5000$

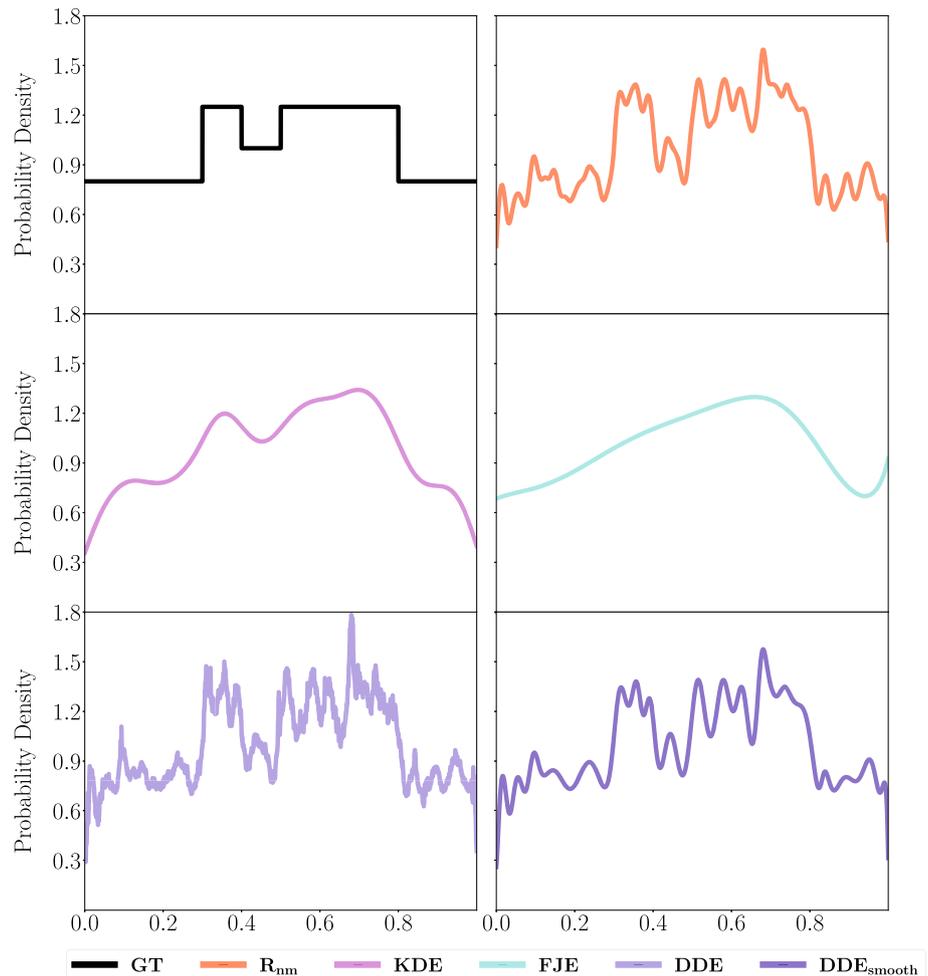

**Author Contributions** T.Ro., T.Ri. and P.H. conceived the initial idea. P.P. T.Ro. and P.H. designed the data generation. P.P. designed the majority of the DDE algorithm with minor contributions from P.H., T.Ri and T.Ro. The experiments were performed by P.P. The results were analyzed by P.P. in collaboration with T.Ro and P.H. The manuscript was written by P.P. in close collaboration with T.Ri. and T.Ro.

**Funding** Open Access funding enabled and organized by Projekt DEAL. This work was partially funded by the Federal Ministry for Economic Affairs and Energy (BMWi) under Grant ZF4483101ED7 (VRReconstruct). We would also like to acknowledge the NVIDIA Corporation for donating a Quadro P6000 for our training cluster.

**Availability of data and materials** The codes for the competing methods and the datasets for the real world samples are publicly available. The proposed codes for DDE and data generation are available on https://github.com/trikpachu/DDE. The complete precomputed datasets used in this paper will be provided after inquiry to P. Puchert.

## Declarations

**Conflict of interest** The authors declare that they have no conflict of interest.

**Code availability** The code for the proposed DDE method, including density estimation, model training and synthetic data generation along with the trained models is available on https://github.com/trikpachu/DDE as well as in the python package deep_density_estimation.